\documentclass[10pt,twocolumn,letterpaper]{article}

\usepackage[pagenumbers]{cvpr} % To force page numbers, e.g. for an arXiv version

\definecolor{cvprblue}{rgb}{0.21,0.49,0.74}
\usepackage[pagebackref,breaklinks,colorlinks,allcolors=cvprblue]{hyperref}
\usepackage{comment}
\usepackage{tcolorbox}
\usepackage[table]{xcolor} 
\newcommand{\codetag}[1]{\texttt{\textcolor{black}{\small#1}}}

\usepackage{booktabs} % For professional quality tables (\toprule, \midrule, \bottomrule)
\usepackage{multirow} % For spanning multiple rows
\usepackage{makecell}   % To have manual line breaks inside a cell
\usepackage{array}      % For the m{} column specifier (vertical centering)

\usepackage[table]{xcolor}
\definecolor{lightblue}{HTML}{E3F2FD}

\usepackage{booktabs,array,colortbl,xcolor}
\newcolumntype{P}[1]{>{\raggedright\arraybackslash}p{#1}}

\usepackage{listings}
\usepackage{tcolorbox}
\usepackage{xcolor}

\usepackage[dvipsnames]{xcolor}
\usepackage{tcolorbox}
\usepackage{listings}
\usepackage{caption}

\definecolor{yamlkeyword}{rgb}{0.5, 0.0, 0.5} % Purple for keywords/keys
\definecolor{yamlvalue}{rgb}{0.0, 0.5, 0.0}   % Green for values (though
\definecolor{yamlnumber}{rgb}{0.1, 0.1, 0.9}   % Blue for numbers
\definecolor{yamlcomment}{rgb}{0.5, 0.5, 0.5} % Gray for comments

\lstdefinestyle{yaml}{
    basicstyle=\footnotesize\ttfamily,
    morekeywords={type, name, size, state, angular_velocity, linear_velocity, 
                  orientation, position, physics, friction, mass, damping, 
                  radius, height, fovy, gravity},
    keywordstyle=\color{yamlkeyword},
    literate=
      *{0}{{{\color{yamlnumber}0}}}1
      {1}{{{\color{yamlnumber}1}}}1
      {2}{{{\color{yamlnumber}2}}}1
      {3}{{{\color{yamlnumber}3}}}1
      {4}{{{\color{yamlnumber}4}}}1
      {5}{{{\color{yamlnumber}5}}}1
      {6}{{{\color{yamlnumber}6}}}1
      {7}{{{\color{yamlnumber}7}}}1
      {8}{{{\color{yamlnumber}8}}}1
      {9}{{{\color{yamlnumber}9}}}1
      {-}{{{\color{yamlnumber}-}}}1
      {.}{{{\color{yamlnumber}.}}}1,
    commentstyle=\itshape\color{yamlcomment},
    breaklines=true,
    keepspaces=true,
    showstringspaces=false,
    frame=none % The tcolorbox provides the frame
}

\usepackage{tabularx} % Required for the new table layout

\newtcbox{\inlineboxBlue}[1][blue!20]{
  on line, 
  colback=blue!3!white,       % Same background   
  colframe=blue!40,           % Same frame color
  left=0.2mm,                   % Same padding as your tcolorbox
  right=0.2mm,                  % Same padding
  top=0.2mm,                  % Same padding
  bottom=0.2mm                % Same padding
}

\newtcbox{\inlineboxRed}[1][red!20]{
  on line, 
  colback=red!3!white,       % Same background   
  colframe=red!40,           % Same frame color
  left=0.2mm,                   % Same padding as your tcolorbox
  right=0.2mm,                  % Same padding
  top=0.2mm,                  % Same padding
  bottom=0.2mm                % Same padding
}

\usepackage{multirow}
\usepackage{times}

\newcommand{\ours}{\textit{$\Delta$}\textsc{ynamics}}
\newcommand{\ourss}{\textit{$\Delta$}\textsc{ynamics}\xspace}

\def\paperID{6659} % *** Enter the Paper ID here
\def\confName{CVPR}
\def\confYear{2026}

\title{\ours: Language-Based Representation for Inferring Rigid-Body Dynamics From Videos}

\author{Chia-Hsiang Kao$^1$\thanks{Corresponding author: ck696@cornell.edu} \quad Cong Phuoc Huynh$^2$ \quad Chien-Yi Wang$^2$ \quad Noranart Vesdapunt$^2$ \\
Stefan Stojanov$^2$ \quad Bharath Hariharan$^1$ \quad Oleksandr Obiednikov$^2$ \quad Ning Zhou$^2$ \\[0.5em]
$^1$Cornell University \quad $^2$Amazon \\[0.3em]
}

\begin{document}
\maketitle

\begin{abstract}
Inferring rigid-body physical states and properties from monocular videos is a fundamental step toward physics-based perception and simulation. 
Existing approaches assume specific underlying physical systems, object types, and camera poses, which are unable to generalize to complex real-world settings. 
We introduce \ours, a vision-language framework that uses language as a unified representation of rigid-body dynamics. Instead of directly predicting parameters, \ourss generates scene configurations in a structured text format for physics simulation.
We enhance the model's generalization by integrating natural language motion reasoning and leveraging optical flow as a semantic-agnostic input. On the CLEVRER dataset~\cite{yi2019clevrer}, \ourss achieves a segmentation IoU of $0.30$, a $7\times$ improvement over leading VLMs (InternVL3-8B, Qwen2.5-VL-7B and Claude-4-Sonnet). Further, test-time sampling and evolutionary search further boost performance by $27\%$ and $120\%$ in segmentation IoU, respectively. 
Finally, we demonstrate strong transfer to a new dataset of $235$ real-world rigid-body videos, highlighting the potential of language-driven physics inference for bridging perception and simulation.
Additional results and videos are available at the project page: \url{https://iandrover.github.io/2026_dynamics/}
\end{abstract} 

\begin{figure}[th]
    \centering
    \begin{subfigure}[b]{0.45\textwidth}
        \centering
        \includegraphics[width=\textwidth]{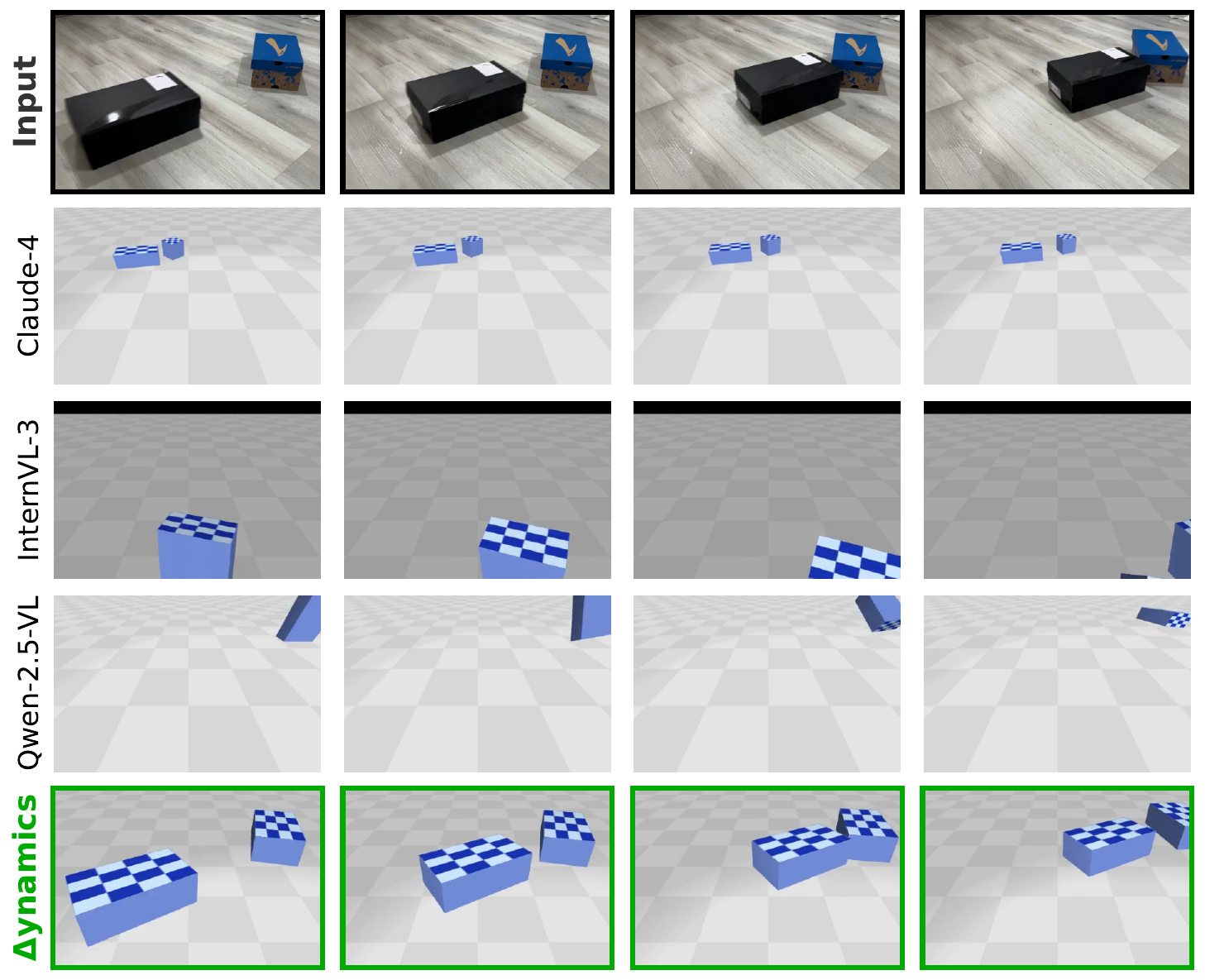}
        \caption{Two shoe boxes collide and slide on the indoor surface.}
    \end{subfigure}
    \begin{subfigure}[b]{0.45\textwidth}
        \centering
        \includegraphics[width=\textwidth]{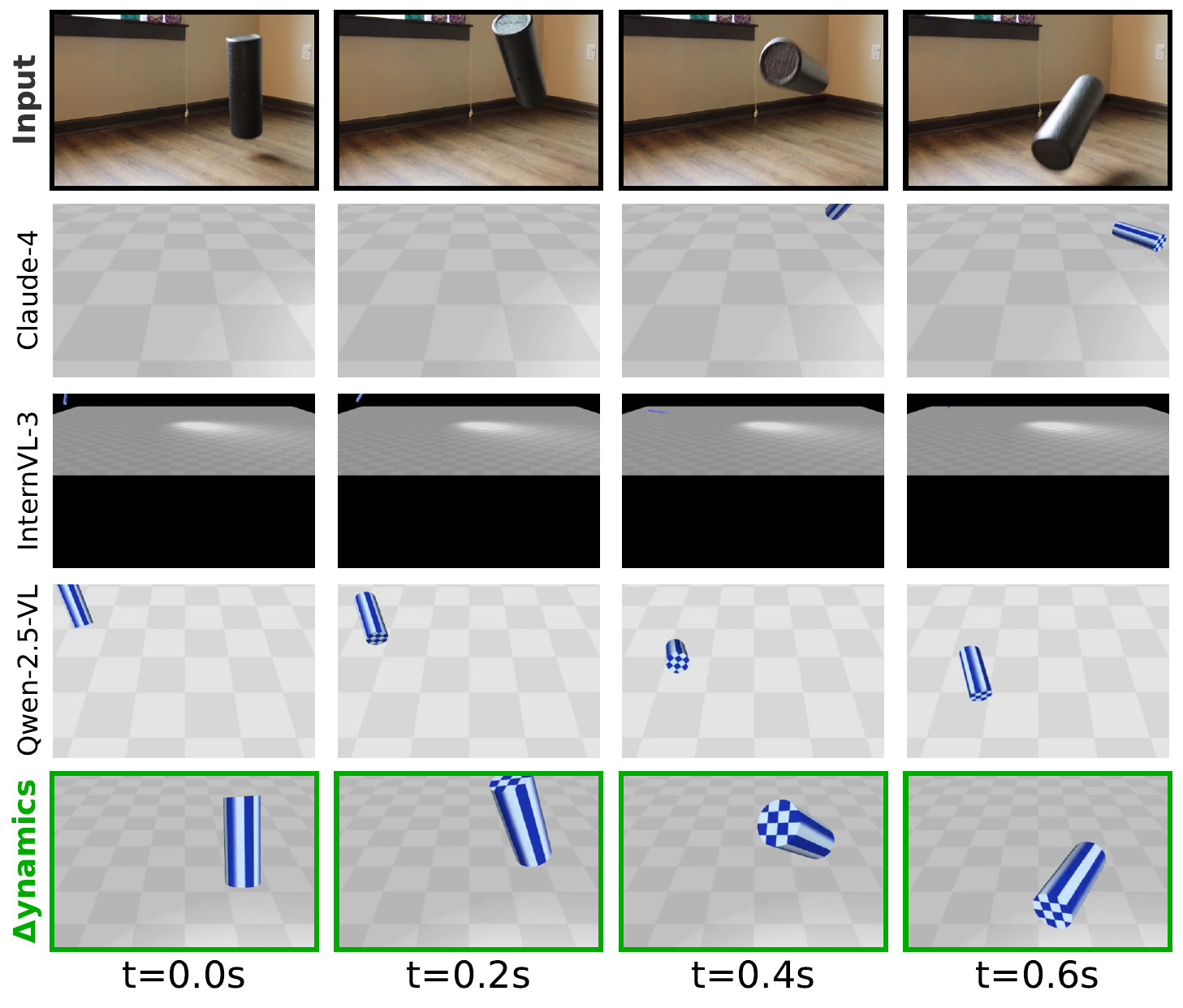}
        \caption{A cylindrical massage roller spins in mid-air.}
    \end{subfigure}
    \vspace{-1mm}
    \caption{\textbf{Motion transfer from real videos to simulation environments.} \ourss accurately reproduces the object shapes, initial position and orientation, material properties, and camera pose with respect to the input videos, while competing VLMs (Claude-4-Sonnet, InternVL-3-8B, Qwen-2.5-VL-7B) fail.}
    \label{fig:teaser}
    \vspace{-5mm}
\end{figure}
\vspace{-2mm}

\section{Introduction}
Understanding physical dynamics from visual observations is a foundational capability for intelligent systems operating in the real world~\citep{battaglia2016interaction, kubricht2017intuitive, watters2017visual, ehrhardt2018unsupervised}. 
When perceiving events such as a ball sliding or bouncing, the system should not only identify and track objects but also infer their intrinsic physical attributes, including friction, elasticity, and other parameters that govern motion.
These inferred properties enable reasoning about cause and effect, anticipating outcomes under varying conditions, and consequently planning and control in embodied settings. 

In this work, we focus on rigid-body motion dynamics.
Given a video, our goal is to infer the underlying physical model that allows the reproduction of the video trajectories within a simulator.
While prior works~\citep{wu2015galileo, wu2017learning, chari2019visual, asenov2019vid2param, heidenlearning, ding2021dynamic, garcia2025learning} have made progress in estimating physics parameters from videos in constrained settings, such as sliding boxes~\cite{wu2015galileo, ding2021dynamic}, billiards~\cite{wu2017learning}, or projectiles~\cite{chari2019visual, asenov2019vid2param, garcia2025learning}, they are not yet applicable to complex real-world motion that involves multiple object interactions and motion types.
First, they assume a model-specific, fixed-length vector of physics parameters for particular object types (e.g., spheres or boxes) and motion types (e.g., sliding or projectile).
This representation is not scalable and does not accommodate the full variety of object interactions. 
Second, prior works typically assume a known or fixed camera pose, thus failing to generalize across varying object distances and camera viewpoints.
As a result, prior methods solve only a narrow subset of this video-to-simulation problem and fail to generalize to complex, real-world scenes involving multiple moving objects and unconstrained viewpoints.

A fundamental challenge lies in how the scene itself is parameterized. In this work, we introduce 
\textbf{\textit{a unified, language-based representation of rigid body motion}} as a bridge between perception and simulation. Instead of regressing a fixed-length numeric vector, we reformulate the problem as the generation of symbolic scene configurations specifying object geometry, initial states, material properties, and camera parameters. 
% These textual configurations are directly executable by physics engines. 
This language  representation is inherently \textit{interpretable} and \textit{scalable} to diverse motion types and object interactions. 

This language-centric formulation naturally motivates the use of Vision-Language Models (VLMs)~\citep{liu2023visual, liu2024improved, deitke2024molmo, chen2024internvl, wu2025qwen}, widely employed for visual reasoning~\citep{antol2015vqa, mathew2021docvqa} and physics understanding~\citep{baradel2019cophy, rajani2020esprit, ates2020craft, riochet2021intphys, chow2025physbench}.
Taking this direction, we develop \ourss, a VLM trained on 400K synthetic videos rendered with MuJoCo~\citep{todorov2012mujoco}, whose output is a YAML format of the scene configuration.
To enhance generalization, we make two key design choices.
First, we take optical flow as the input as it is agnostic to visual semantics and background, which provides explicit motion cues and improves full-sequence segmentation IoU by $26\%$ (from 0.19 to 0.24) on CLEVRER~\citep{yi2019clevrer}. 
Second, we augment supervision with \textit{natural-language motion descriptions} that capture trajectories, object visibility, and collision events as an auxiliary textual target.
Together, these components make \ourss robust to domain shifts. 

To evaluate cross-domain generalization, we adapt CLEVRER for controlled testing and curate a new dataset of 235 real-world rigid-body motion videos. The reasoning-enhanced model consistently outperforms the vanilla version on real-world transfer, indicating stronger generalization capabilities.
We also investigate several test-time enhancement strategies that do not require labeled ground truth in the target domain. We find that best-of-k sampling consistently yields a 10\% improvement, and that an additional evolutionary search provides over 50\% further gains.
For real-world application, we also show the potential of physically plausible video editing using our framework; the corresponding results are deferred to Appendix~\ref{app:editing}.

Our main contributions are summarized below:
\begin{itemize}
    \item \textbf{Language-based representation for motion dynamics:}
    We reformulate rigid object motion estimation from videos as a \textit{language modeling} problem, where the model generates structural textual scene configurations that are directly consumable by a physics engine. 
    \item \textbf{VLM for rigid-body physics inference:}
    We present \ours, a VLM that directly infers the underlying physics parameters of rigid object motion, which enables the reconstruction of physically-plausible motion trajectories from monocular videos.

    \item \textbf{Cross-domain generalization:}
    We boost the generalization of models for different physics engines and real-world videos by introducing two key innovations: using optical flow as semantics-agnostic input and training the model to predict natural-language motion descriptions.

    \item \textbf{Comprehensive evaluation benchmarks:}
    We adapt the CLEVRER dataset for controlled benchmarking and curate a new dataset of $235$ real-world rigid-body motion videos with corresponding annotations for segmentation masks and optical flows. 
\end{itemize}
\begin{figure*}[t]
\centering
\includegraphics[width=0.98\textwidth]{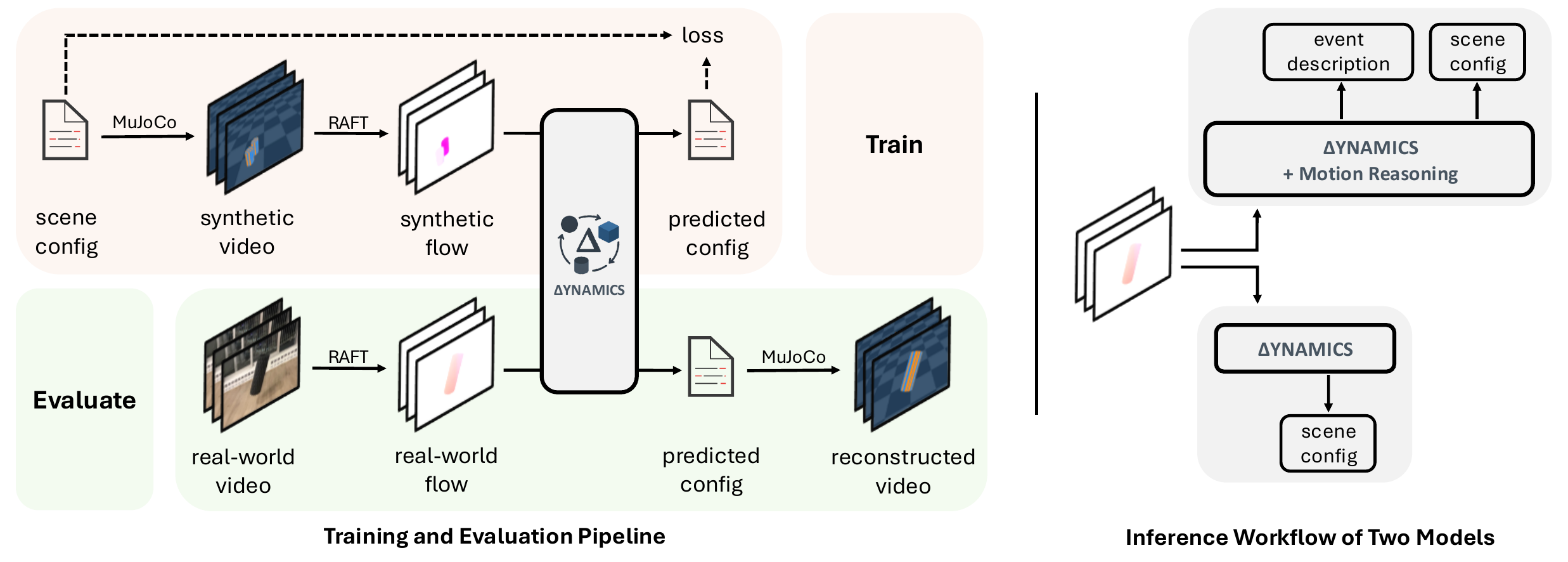}
\vspace{-2mm}
\caption{
\textbf{Training, evaluation and inference workflow for \ours~.}
\textbf{Training (top left):} We sample scene configurations and render corresponding synthetic videos using the MuJoCo physics engine. Next, we compute optical flows using RAFT~\citep{teed2020raft} and train \ourss to generate scene configurations in a structured text format given optical flows.  
\textbf{Evaluation (bottom left):} \ours~takes input optical flows derived from real-world videos to infer scene configurations. \textbf{Inference (right):} The base variant (bottom) directly generates the scene configuration, whereas the motion reasoning variant (top) first generates a motion event description (details illustrated in Figure~\ref{fig:reasoning}), then predicts the scene configuration.
}
\vspace{-2mm}
\label{fig:pipeline}
\end{figure*}

\section{Related Work}

\noindent\textbf{Rigid-Body Motion Parameter Estimation.}
Estimating physics parameters of rigid moving objects and camera geometry from videos is a key step towards physics-based perception and simulation. Early efforts tackled the problem in narrow scenarios, e.g., sliding boxes~\cite{wu2015galileo, ding2021dynamic}, billiard games~\cite{wu2017learning}, projectile motion~\citep{chari2019visual, asenov2019vid2param}, articulated rigid body~\cite{heidenlearning}, or free fall~\citep{garcia2025learning}), to keep the parameter estimation problem tractable. Furthermore, they assume fixed camera parameters, preventing their applications from general real-world settings~\citep{wu2015galileo, wu2017learning, asenov2019vid2param, chari2019visual, heidenlearning, ding2021dynamic, garcia2025learning}. Our contribution is a general solution to the physics parameter estimation problem, which is applicable to a wide spectrum of physical motions in unconstrained real-world settings. 

\vspace{0.4em}
\noindent\textbf{Structured Representations for Visual Content.}
Representing image and video content using structured graphics programs has been widely adopted for graphics simulation engines such as Blender and MuJoCo format~\cite{todorov2012mujoco}. 
Recent works in image simulation and generation make use of programmatic formats such as SVG~\cite{xing2024svgdreamer, song2025layertracer, rodriguez2025starvector, xing2025svgdreamer++, wu2025chat2svg, chen2025symbolic} and TikZ~\cite{belouadi2024detikzify, belouadi2025tikzero, tan2025sketchagent} to formulate the problem as conditional generation of structured text based on textual and image prompts using diffusion models.
Further, the use of structured graphics programs has also been extended to the domain of inverse rendering and the editing and generation of 3D scenes. Specifically, the workflow in~\cite{kulits2024re, bian2025chatgarment, kulits2025reconstructing, zhang2025scene} involves training VLMs to translate images into a structured format (e.g., JSON) and employing graphic engines for rendering. Other works train VLMs to infer the programmatic representation of existing scenes for graphics editing~\cite{gu2025blendergym} and 3D asset creation~\citep{zhao2025di}. Our proposed approach is inspired by this school of thought because it offers the capability to interpret both the underlying physics and controllable editing of visual content via scene attributes. However, our research problem is distinguished from prior works by the focus on the modeling of motion dynamics in videos. 

\vspace{0.4em}
\noindent\textbf{Physics Simulation.}
Another line of research focuses on end-to-end training with differentiable simulation and rendering pipelines~\cite{hu2019difftaichi, huang2021plasticinelab, freeman2021brax}, enabling physical scene understanding via gradient-based optimization. These works~\cite{jatavallabhula2021gradsim, yang2023ppr, sundaresan2022diffcloud, qiao2022neuphysics, mittal2025uniphy} jointly optimize object geometry and physical parameters to directly match the scene dynamics and image formation.
Our approach differs in three ways. First, we do not require a predefined physics model; instead, our model learns to infer physical properties and dynamics directly from video observations. Second, we do not assume access to differentiable simulators, as most physics engines are not differentiable in practice. Third, instead of per-scene optimization, we employ a direct feedforward model that predicts a complete scene configuration in a single pass.
\section{Method}
We present the training, evaluation and inference workflows of~\ours~in Figure~\ref{fig:pipeline}. 
We now formalize the problem and describe each component in detail.
% The left panel shows the training and evaluation pipelines. During training, we synthesize videos from sampled scene configurations, derive optical flows from these synthetic videos, and train~\ours~to predict configurations (physics parameters and camera geometry) from the input optical flows. 
% For evaluation, we use the MuJoCo physics engine to render a video from the predicted scene configuration and assess motion motion similarity between the reconstructed and original input video. 
% The right panel depicts inference workflows for two model variants: the base variant (bottom) predicts scene configurations only, while the enhanced variant (top) additionally generates natural language descriptions of the observed motion.

\subsection{Problem Statement}
We address the problem of recovering physical scene parameters and dynamics from a monocular video.
Given an input video $\mathbf{X}$, a model $\mathcal{F}_\theta$ predicts a parameter set $\mathbf{c} = \mathcal{F}_\theta(\mathbf{X})$, which is then provided to a physics engine $\mathcal{S}$ to generate a reconstructed sequence $\hat{\mathbf{X}} = \mathcal{S}(\mathbf{c})$.
The objective is to learn $\mathcal{F}_\theta$ such that the simulated dynamics in $\hat{\mathbf{X}}$ faithfully reproduce those observed in the input video.

\subsection{Unified Scene Representation}\label{sec:method_scene_representation}

A fundamental challenge lies in how the scene itself is parameterized. 
Prior work typically regresses a fixed-length parameter vector specific to a particular object set, simulation model, or physics system, which limits generalization across real-world scenarios.

\vspace{0.4em}
\noindent \textbf{Language Representation.}
To address these limitations, we shift the core paradigm from numerical regression to symbolic generation. 
The key idea is a unified, language-based representation that acts as a bridge between perception and simulation. 
Specifically, we recast physics estimation as a text-generation problem: the model outputs a YAML-formatted sequence that encodes the entire scene configuration, including object geometry, initial states, material properties, and camera parameters.
This provides three key advantages:
\vspace{0.2em}
\begin{itemize}
    \item \textbf{Extensibility and Interpretability.}
A textual format scales naturally to scenes with arbitrary numbers of objects. It is human-readable, easy to edit, and conducive to counterfactual analysis. Extended results on physically plausible video editing are provided in Appendix~\ref{app:editing}.
    \item \textbf{Natural Integration with VLM.}  
Casting simulation as text generation enables end-to-end training of a unified VLM without engineering multi-stage components.  
We simply format the target as \codetag{<answer> {configuration} </answer>}, allowing the model to directly output the scene description.
    \item \textbf{Joint Reasoning and Configuration Generation.}  
Language models can interleave descriptive reasoning with configuration prediction, enabling richer intermediate representations within the same autoregressive process.
\end{itemize}

\begin{table}[t]
\centering
\footnotesize
\setlength{\tabcolsep}{6pt}
\renewcommand{\arraystretch}{1.05}
\caption{\textbf{Parameter categories.} The complete parameter space spans object properties, initial states, and global parameters.}
\vspace{-3mm}
\begin{tabular}{m{0.13\linewidth} p{0.75\linewidth}}
\toprule
\textbf{Category} & \textbf{Parameters} \\ 
\midrule
\multirow{2}{*}{\textbf{\makecell[l]{Object \\ Property}}} & 
\textit{Geometry / Inertial:} radius, height, width, depth, mass. \\
& \textit{Material:} friction (rolling, sliding) and damping. \\[3pt]

\multirow{2}{*}{\textbf{\makecell[l]{Initial \\ State}}} & 
\textit{Kinematics:} position, linear and angular velocity. \\
& \textit{Orientation:} quaternion. \\[3pt]

\multirow{2}{*}{\textbf{\makecell[l]{Global \\ Parameter}}} & 
\textit{Camera:} pose (height, angle, FOV). \\
& \textit{Environment:} gravity. \\
\bottomrule
\label{tab:scene_config}
\end{tabular}
\end{table}

\vspace{0.4em}
\noindent \textbf{Parameterization Details.}
To operationalize this language-based approach, we define a structured schema for the scene configuration. This configuration represents the full set of geometry, physics, and camera parameters required for simulation, as summarized in Table~\ref{tab:scene_config}. 
We compose our scenes using three primitive shapes (spheres, cylinders, and boxes) that can cover common household objects such as tennis balls, soda cans, mugs, books, and crates. 
These primitives are sufficient to simulate essential rigid-body dynamics, including bouncing, rolling, sliding, and collisions.
For the camera, we place it at $(0, -2, h)$, where $h$ denotes its height, and vary the pitch angle while setting roll and yaw to zero.
We include gravity as a parameter to account for variations in frame rate or time scale.
This scene configuration format supports both single-object and arbitrary multi-object scenes.
For example, a scene containing four box-shaped objects includes $20 \times 4$ box-specific parameters, along with $3$ camera parameters and $1$ gravity term, totaling $84$ parameters to be estimated.

\subsection{Motion Reasoning}
\label{subsec:method_motion_reasoning}

\begin{figure}[t]
\centering
\includegraphics[width=\columnwidth]{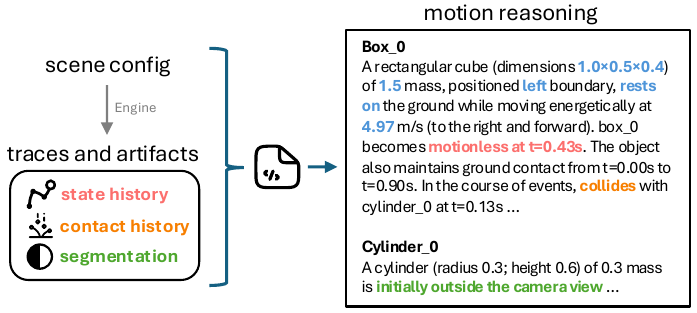}
\vspace{-6mm}
\caption{\textbf{Synthetic training data generation.} 
During the data generation process, we create natural language descriptions of motion events. An event-mining script processes the simulation traces and artifacts (left), including state history, contact history, and segmentation maps, to find key dynamic events. The resulting textual descriptions (right) serve as ground-truth targets for the motion reasoning model during training.}
\label{fig:reasoning}
\end{figure}

By reasoning about motion and object interactions before predicting scene parameters, the model learn richer representations of the underlying dynamics, which in turn improve the accuracy of the subsequent scene parameter estimates. 
To enable motion reasoning, we train a variant of the model that first generates a natural language description of the observed dynamics and then produces the scene configuration. As shown in Figure~\ref{fig:reasoning}, these descriptions are derived from simulation traces and artifacts (i.e., object state histories, contact logs, segmentation masks) together with ground-truth configurations. We specifically consider events such as visibility (e.g., when the object enters or leaves the camera view), motion change (e.g., when it stops rolling or sliding), and collisions (e.g., when it touches the ground or another object), and we design rule-based functions to parse them. These detected events are then inserted into predefined templates to generate a structured, natural language description of the motion.
Finally, we prepend the motion description to the scene configuration as \codetag{<think> {description} </think> <answer> {configuration} </answer>}. Table~\ref{tab:reasoning_appendix_example} provides an illustrative example.

\subsection{Motion-Aware Input Representation}
Raw RGB videos contain visual semantics unrelated to motion, which can introduce confounding factors for our model.
As an alternative, we use optical flow fields to represent motion as it is agnostic to visual semantics and appearance. Specifically, we compute optical flows using RAFT~\citep{teed2020raft} and convert them into a 2D array per color channel, which can further be fed into VLM without architectural changes. We evaluate models trained with both  RGB video inputs and the RGB-transformed optical flow maps.
\section{Training and Evaluation Method}

\subsection{Training Approach}\label{subsec:method_trainin_data}

% In our problem setting, we need to deal with a given physics simulator which is often differentiable. Therefore, our problem is focused on learning the scene configuration predictor, instead of learning both the configuration predictor and physics engine end-to-end. 

\noindent \textbf{Synthetic Data Curation.}
We generate a dataset of 400K unique physical scenes using the MuJoCo simulator~\citep{todorov2012mujoco}. 
Each training example is created by sampling a full YAML scene configuration, which is then converted into MuJoCo’s XML format to initialize the simulator and assign dynamic states (e.g., initial velocities), as detailed in Appendix~\ref{app:yaml_xml}.
Each data point includes a rendered RGB video of a scene with up to four objects and the corresponding YAML file. 
To specifically test for compositional generalization, four distinct object-type combinations in four-object scenes (e.g., two boxes and two cylinders) are also held out.

To ensure physically plausible and visually meaningful interactions, we filter out (i) scenes with overlapping objects at initialization, identified using MuJoCo's built-in collision detection; (ii) scenes where more than one  object remains outside the camera’s field of view throughout simulation; and (iii) scenes where any object is too small (with a total area less than $8000$ pixels).
Simulations are run for 1~second at 30~FPS with eight physics substeps per frame, rendering images at $480 {\times} 320$ resolution.

\vspace{0.4em}
\noindent \textbf{Learning Objective.}
Let $\mathcal{D} = \{(\mathbf{X}_i, \mathbf{c}_i)\}_{i=1}^N$ denote the training set, where $\mathbf{X}_i$ is a video observation and $\mathbf{c}_i$ is the corresponding scene configuration, represented as a tokenized text sequence.
We aim to learn a model $\mathcal{F}_\theta$ that maximizes the conditional likelihood $p_\theta(\mathbf{c} \mid \mathbf{X})$.
This likelihood is modeled autoregressively over the tokens of $\mathbf{c}$ as
\begin{equation}
    p_\theta(\mathbf{c} \mid \mathbf{X})
    = \prod_{t=1}^{|\mathbf{c}|}
      p_\theta(c_t \mid \mathbf{X}, c_{<t}),
\end{equation}
where $c_t$ denotes the $t$-th token of $\mathbf{c}$. We train $\mathcal{F}_\theta$ by minimizing the negative log-likelihood (NLL) over the dataset:
\begin{equation}
    \mathcal{L}_{\text{VLM}}
    = -\sum_{(\mathbf{X}, \mathbf{c}) \in \mathcal{D}}
      \log p_\theta(\mathbf{c} \mid \mathbf{X}).
\end{equation}

\vspace{0.4em}
\noindent \textbf{Model and Training Implementation.}
Our architecture is based on Qwen2.5-VL-3B~\citep{bai2025qwen2}. 
The inputs consist of $10$ frames uniformly sampled from 1-second, 30~FPS videos. 
We train two variants, one predicting only scene configuration, and the other generating motion reasoning as an additional output. The format of the target text sequences for these variants has been described in Sections~\ref{sec:method_scene_representation} and ~\ref{subsec:method_motion_reasoning}.
We fine-tune the full model for 10~epochs in bfloat16 mixed precision on eight 40~GB A100 GPUs. We employ the AdamW optimizer with a learning rate of $2{\times}10^{-5}$, weight decay of $0.01$, and a global batch size of~$128$.

\subsection{Test-Time Strategy} \label{subsec:method_preference_optimization}

To improve instance-level accuracy at inference time, we explore three complementary test-time optimization strategies: best-of-K sampling, preference-based refinement, and evolutionary search.

\vspace{0.4em}
\noindent \textbf{Best-of-K Sampling.}
The greedy decoding strategy in VLMs is not guaranteed to find the parameter set with the best quality, as the optimal parameter set may lie in the long tail of the model's output distribution. 
Hence, we adopt a best-of-$N$ evaluation scheme to explore this distribution: for each case, we generate $N=32$ diverse predictions with a temperature of $0.1$ and top-p of $0.9$ and report the \textit{Best@32} performance. This reflects the model’s ability to recover accurate physical dynamics with multiple attempts.

\vspace{0.4em}
\noindent \textbf{Preference Optimization.}
While ground-truth configurations are typically unavailable for novel environments, the similarities between the forward rendering and the input videos, such as object mask IoU, can serve as \emph{implicit reward signals} for configuration selection without explicit supervision. We conduct experiments with preference rank optimization~\citep{song2024preference}, where details and relevant results are provided in Appendix~\ref{app:method_preference_optimization}.

\vspace{0.4em}
\noindent \textbf{Evolutionary Search.}
Since preference optimization generally requires training data, it is generally impractical in real-world scenarios. Thus, we explore Covariance Matrix Adaptation Evolution Strategy (CMA-ES)~\citep{hansen1996adapting}, which is an evolutionary algorithm suited for non-convex black-box optimization. When using CMA-ES, we initialize the search with the \textit{Best@32} sample, and we optimize scene 
configurations, including object sizes, initial states, physics parameters, and camera poses, while keeping object types fixed. We employ a heuristic fitness function that maximizes the segmentation Intersection-over-Union (IoU) while minimizing the optical flow end-point error (EPE), formulated as $(\text{IoU} - \text{EPE})$.  
We use a population size of $128$ and optimize for $100$ iterations.

\subsection{Evaluation Method}\label{subsec:method_eval}
\noindent\textbf{Evaluation via Simulation.}
We evaluate the quality of the predicted configuration $\hat{\mathbf{c}}$ through re-simulation. 
The generated text is passed to the physics engine $\mathcal{S}$ to produce a simulated RGB video $\hat{\mathbf{X}} = \mathcal{S}(\hat{\mathbf{c}})$, along with auxiliary outputs such as object segmentation masks and optical flow fields.
We then compare the simulated outputs against the ground-truth counterparts from $\mathbf{X}$ using segmentation Intersection-over-Union (IoU) and optical flow end-point error (EPE). For synthetic data, ground-truth masks are available from the renderer, while for real-world videos, we use pretrained models~\citep{ravi2024sam, teed2020raft} for pseudo-annotation.

\vspace{0.4em}
\noindent \textbf{Metrics.} We evaluate across three dimensions:
\begin{itemize}
    \item {Object Composition}: Accuracy of object composition.
    \item {Motion Reconstruction Quality}: Similarity between the re-simulated and reference videos, measured by segmentation IoU and flow EPE.
    \item {Physics Parameter}: $L_1$ distance between estimated and ground-truth parameters, computed only when the object combination is correct.
\end{itemize}

\vspace{0.4em}

\noindent\textbf{Baselines.}
Since our work is the first to estimate a complete scene configuration for diverse physical systems from a single monocular video, it is not directly comparable to prior methods on physics parameter estimation. 
For evaluation, we establish baselines using both proprietary and open-source vision–language models (VLMs), including InternVL3-8B~\citep{zhu2025internvl3}, Qwen2.5-VL-7B~\citep{bai2025qwen2}, and Claude-4-Sonnet~\citep{claude}.
Each model is evaluated using \emph{three-shot in-context learning (ICL)}. 
We adopt ICL over zero-shot prompting because  
(1) zero-shot generation of a MuJoCo XML file is insufficient for running a simulation, since dynamic states can only be set after engine initialization; and   
(2) Generating both the XML file and the dynamic-state initialization code is difficult. Details about a few-shot examples are deferred to Appendix~\ref{app:icl_examples}.

We also establish non-VLM baselines that directly predict scene configuration parameters from videos. 
We concatenate these parameters into a fixed-length vector, represent the object type with one-hot encoding, and perform zero-padding for missing objects. Objects are ordered by their $x$- and then $y$-positions. We adopt the pretrained ViViT~\citep{arnab2021vivit} model, designed for video classification, and fully fine-tune it using an $\ell_2$ loss on the regression targets.

\begin{table*}[thbp]
\centering
\scriptsize
\caption{\textbf{Evaluation metrics for the in-distribution setting on the synthetic evaluation data.}
% We show object composition accuracy, motion reconstruction accuracy (segmentation map IoU and optical flow EPE), and physics parameter estimation accuracy. 
When \ours~ takes optical flows as the input, it consistently outperforms baseline methods across most evaluation dimensions. Note that parameter estimation metrics are unavailable for non-VLM baselines since they do not correctly predict the object composition. 
% Best values are in \textbf{bold}.
\textbf{Best} and \underline{runner-up} results are highlighted.
}
\vspace{-2mm}
\begin{tabular}{l c|c|cc|cc|ccc}
\toprule
 & & \textbf{Obj. Comp.} & \multicolumn{2}{c|}{\textbf{Segmentation Map IoU (↑)}} & \multicolumn{2}{c|}{\textbf{Optical Flow EPE (↓)}} & \multicolumn{3}{c}{\textbf{Physics Parameter MAE (↓)}} \\
 & Input & \textbf{Acc. (↑)} & First-Frame & Full Sequence & First-Frame & Full Sequence & Damping & Roll Friction & Slide Friction \\
\midrule
\rowcolor{lightblue} \multicolumn{10}{l}{\textbf{\textit{Non-VLM Models}}} \\
ViViT~\cite{arnab2021vivit} & RGB & 0.00 & 0.08 & 0.07 & 18.52 & 9.38 & -- & -- & -- \\
ViViT~\cite{arnab2021vivit} & Opt. Flow & 0.00 & 0.07 & 0.06 & 8.54 & 8.90 & -- & -- & -- \\
\midrule
\rowcolor{lightblue} \multicolumn{10}{l}{\textbf{\textit{VLM Models}}} \\
InternVL3-8B~\cite{zhu2025internvl3} & RGB & 0.02 & 0.05 & 0.05 & 25.13 & 15.77 & 2.94 & 0.35 & 0.81 \\
Qwen2.5-VL-7B~\cite{bai2025qwen2} & RGB & 0.27 & 0.03 & 0.03 & 39.98 & 16.33 & 1.97 & 0.32 & 0.66 \\
Claude-4-Sonnet~\cite{claude} & RGB & 0.45 & 0.09 & 0.07 & 13.79 & 11.07 & 1.71 & 0.24 & 0.43 \\
\midrule
\rowcolor{lightblue} \multicolumn{10}{l}{\textbf{\textit{Ours}}} \\
\ours & RGB & 0.60 & 0.52 & 0.32 & 27.58 & 19.66 & \textbf{1.52} & 0.16 & 0.16 \\
\ours & Opt. Flow & 0.97 & 0.88 & 0.49 & \phantom{0}5.75 & \phantom{0}9.24 & 1.72 & \textbf{0.15} & 0.16 \\
+ Motion Reasoning & Opt. Flow & \textbf{0.99} & \textbf{0.91} & \textbf{0.54} & \textbf{\phantom{0}4.88} & \textbf{\phantom{0}8.52} & 1.60 & 0.16 & \textbf{0.15} \\
\bottomrule
\end{tabular}
\label{tab:eval_mujoco_simplified}
\end{table*}

\begin{table}[thbp]
\centering
\scriptsize
\caption{
\textbf{Robustness to complex scene dynamics.} \ourss models, trained on up to four objects, generalize effectively to more complex scenes with up to six interacting objects. Structured motion reasoning enhances robustness and consistency under increasing scene complexity. Note that four four-object configurations are held out during training and evaluated here to assess true out-of-distribution generalization.
}
\vspace{-2mm}
\begin{tabular}{ll cc}
\toprule
& & \multicolumn{2}{c}{\textbf{Segmentation Map IoU (↑)}}\\
\# Objects & Model & First Frame & Full Sequence\\
\midrule
4 & \ours &  0.88 & 0.53 \\
 & + Motion Reasoning & 0.89 & 0.54 \\
\midrule
5 & \ours & 0.87 & 0.51  \\
 & + Motion Reasoning & 0.88 & 0.54 \\
\midrule
6 & \ours & 0.85 & 0.50  \\
 & + Motion Reasoning & 0.81 & 0.52 \\
\bottomrule
\end{tabular}
\label{tab:eval_mujoco_ood_simplified}
\end{table}

\section{Results on Synthetic Dataset}
% \subsection{Setup}
We divide the synthetic data into two subsets and  evaluate the model in the following settings
\begin{enumerate}
    \item {Comparative evaluation}: Scenes containing 1--3 objects, with 100 samples for each object count. 
    \item {Complex scene dynamics}: 400 four-object scenes constructed from the four specific object-type combinations that were excluded from the training set. 400 scenes with five objects and 400 scenes with six objects, which exceed the object counts seen during training.
\end{enumerate}

\subsection{Comparison with Baselines}
As shown in Table~\ref{tab:eval_mujoco_simplified},  Claude-4 is the best model among the baselines. While sufficient to roughly identify object composition in a scene, they perform poorly in reconstructing motion trajectories and estimating the physics parameters, with low segmentation IoU ($ \leq 0.09$), and high optical flow errors ($ > 11$) observed.

% Meanwhile, the base model of ~\ours, which takes RGB videos as the input, outperforms the baselines in most metrics, except the optical flow end-point error (EPE). This could be because the scene configuration generation is not directly guided by optical flows as a supervisory signal. 
Meanwhile, the RGB-based \ours\ model outperforms all baselines in object composition accuracy, segmentation IoU, and parameter estimation, but underperforms in optical flow end-point error (EPE). 
The higher EPE is primarily due to occasional interpenetration in the predicted initial states, which causes MuJoCo to apply large corrective contact forces to separate overlapping objects, resulting in abrupt motions that increase flow error.

When explicitly conditioned on optical flow, \ourss achieves 97\% object composition accuracy, improves segmentation IoU, and substantially reduces EPE. One exception is the damping estimation, where the raw-RGB model performs slightly better, likely because the checkerboard ground plane provides additional visual cues helpful to estimating damping parameters. Finally, adding motion reasoning, as shown in the last row, further improves overall performance.

\subsection{Evaluation on Complex Scenes}
Next, we evaluate the model's generalization ability on unseen scene configurations. In particular, we focus on the behavior of \ours\ variant with motion reasoning (which is the best one based on the previous section). 
In Table~\ref{tab:eval_mujoco_ood_simplified}, the model shows only a marginal degradation of segmentation map IoU (compared to the last row of Table~\ref{tab:eval_mujoco_simplified}) for scenes with  4 or 5 objects. Even for scenes with 6 objects, the degradation is gradual and slow. 
These results show that incorporating motion reasoning adds robustness to more complex, unseen multi-object dynamics. 

% \vspace{-5mm}
\begin{table}[thbp]
\centering
\scriptsize
\caption{\textbf{Cross-engine generalization.} Evaluating transfer from MuJoCo (training) to Blender (CLEVRER~\cite{yi2019clevrer}) demonstrates that \ours~maintains its performance in a zero-shot setting despite domain shifts. Incorporating structured motion description consistently improves segmentation map IoU. 
}
\label{tab:eval_clevrer_baseline}
\vspace{-2mm}
\begin{tabular}{l c|cc}
\toprule
 & & \multicolumn{2}{c}{\textbf{Segmentation Map IoU (↑)}} \\
 & Modality & First Frame & Full Sequence \\
\midrule
\rowcolor{lightblue} \multicolumn{4}{l}{\textit{\textbf{VLM Models}}} \\
InternVL3-8B & RGB & 0.01 & 0.02 \\
Qwen2.5-VL-7B & RGB & 0.01 & 0.01  \\
Claude-4-Sonnet & RGB & 0.03 & 0.04  \\
\midrule
\rowcolor{lightblue} \multicolumn{4}{l}{\textit{\textbf{Ours}}} \\
\ours & RGB & 0.43 & 0.19 \\
\ours & Opt. Flow & \underline{0.63} & \underline{0.24}  \\
+ Motion Reasoning & Opt. Flow & \textbf{0.67} & \textbf{0.30} \\
% + + PRO & OF & \underline{0.68} & 0.31 \\
% + + best@32 & OF & \textbf{0.76} & \underline{0.38} \\
% + + CMA-ES & OF & 0.63 & \textbf{0.64} \\
\bottomrule
\end{tabular}
\end{table}

\begin{table*}[thbp]
\centering
\scriptsize
\caption{\textbf{Evaluation of test-time optimization strategies on CLEVRER.}
We compare the base and motion reasoning variants of \ourss with greedy decoding, best-of-$32$ sampling under temperature of 0.1, and evolutionary search (CMA-ES). \textit{Best@1} denotes the average of the 32 samples, while \textit{Best@32} reports the best. 
}
\vspace{-2mm}
\begin{tabular}{l l l c|ccc|ccc|ccc}
\toprule
& \multicolumn{6}{c}{\textbf{Segmentation Map IoU ($\uparrow$)}} & \multicolumn{6}{c}{\textbf{Optical Flow EPE ($\downarrow$)}} \\
\cmidrule{2-13}
& \multicolumn{3}{c}{First Frame} & \multicolumn{3}{c}{Full Sequence} & \multicolumn{3}{c}{First Frame} & \multicolumn{3}{c}{Full Sequence} \\
& Greedy & Best@1 & Best@32 & Greedy & Best@1 & Best@32 & Greedy & Best@1 & Best@32 & Greedy & Best@1 & Best@32 \\
\midrule
\multirow{1}{*}{\ours} & 0.63 & 0.63 & 0.67 & 0.24 & 0.24 & 0.28 & 3.66 & 3.65 & 2.92 & 6.91 & 6.86 & 6.21 \\
% \midrule
% \cmidrule{2-13}
\multirow{1}{*}{+ Motion Reasoning} 
& {0.67} & \underline{0.68} & {\textbf{0.76}} 
& {0.30} & {0.30} & \underline{0.38}
& {2.92} & {2.93} & \underline{2.22} 
& {5.94} & {5.95} & \underline{5.17} \\
% \midrule
% \cmidrule{2-13}
% \multirow{1}{*}{+ + PRO}
% & {0.68} & \underline{0.69} & \textbf{0.77} 
% & {0.31} & {0.31} & \underline{0.39} 
% & {2.90} & {2.94} & \underline{1.85} 
% & {5.78} & {5.81} & \underline{4.78} \\
\multirow{1}{*}{+ + CMA-ES}
& {0.62} & - & - 
& \textbf{0.66} & - & - 
& \textbf{0.13} & - & - 
& \textbf{0.11} & - & - \\
\bottomrule
\end{tabular}
\label{tab:eval_CLEVRER_sampling_rl_simplifed}
\end{table*}

\section{Cross-Engine and Real-World Results}

\subsection{Cross-Engine Generalization}

We assess our model's ability to generalize across a fundamentally different simulation and rendering engine. In particular, we perform this evaluation on the CLEVRER dataset~\cite{yi2019clevrer}. CLEVRER is a video question-answering benchmark rendered by Blender, which covers sliding motion dynamics for three object types: cubes, spheres, and cylinders. 
We sampled 100 test videos from the CLEVRER for evaluation. To evaluate against the baseline VLMs, we adopt a similar few-shot, in-context prompting approach (details in Appendix~\ref{app:icl_examples}).

% Results in Table~\ref{tab:eval_clevrer_baseline} show that~\ourss successfully captures multi-object motion in this new domain. Although overall metric values are lower, partly due to CLEVRER’s relatively small object scale—the same trends hold: models using optical flow inputs outperform RGB-based ones, and incorporating reasoning further increases performance, e.g., full-sequence segmentation map IoU from 0.24 to 0.29 (a 21\% improvement). 

\begin{figure}[t]
    \centering
    \begin{subfigure}[b]{0.45\textwidth}
        \centering
        \includegraphics[width=\linewidth]{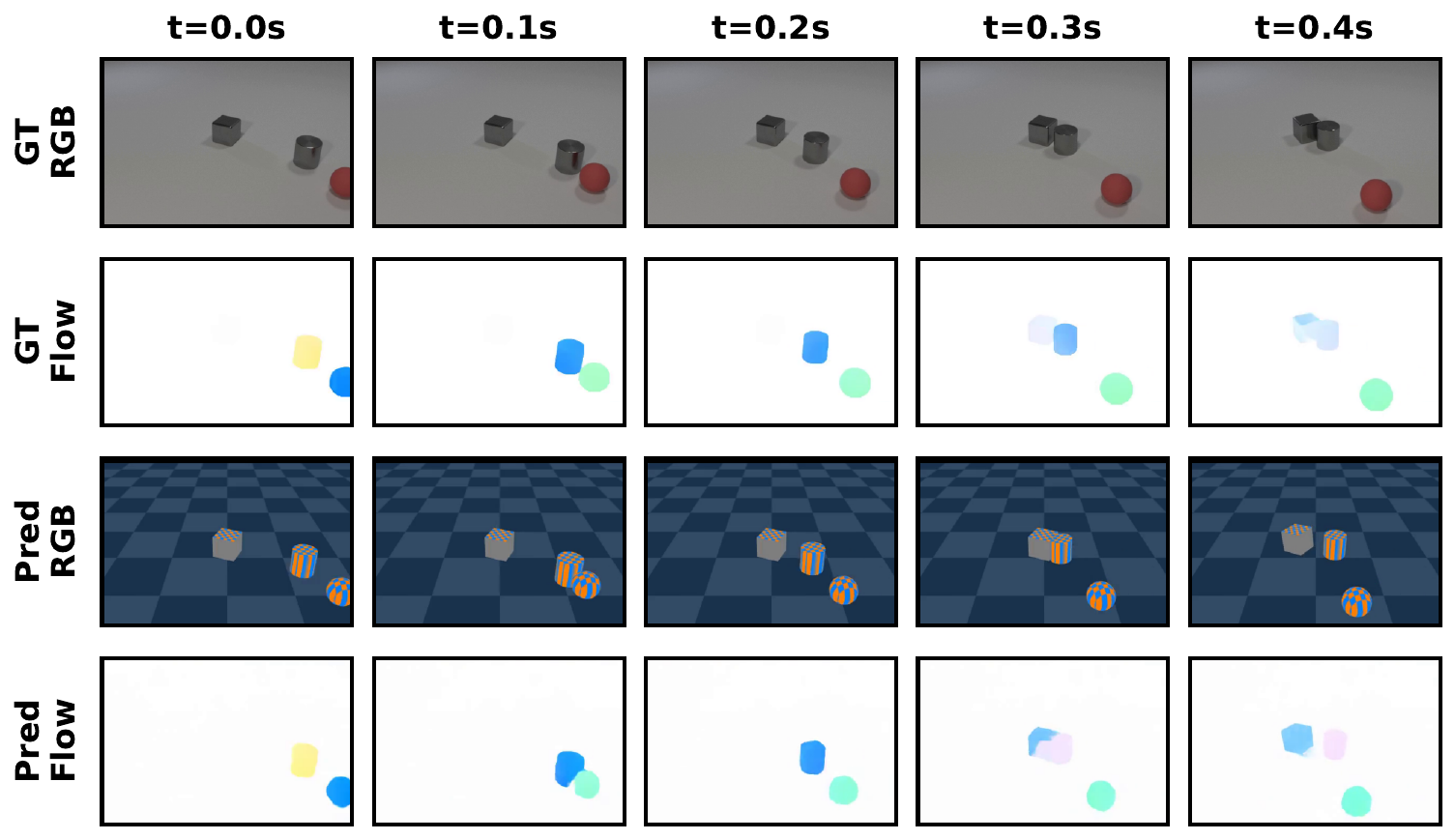}
        \caption{A moving box collides with a static cube; a red ball rolls nearby.}
        \label{fig:eval_real_world_2}
    \end{subfigure}
    \begin{subfigure}[b]{0.45\textwidth}
        \centering
        \includegraphics[width=\linewidth]{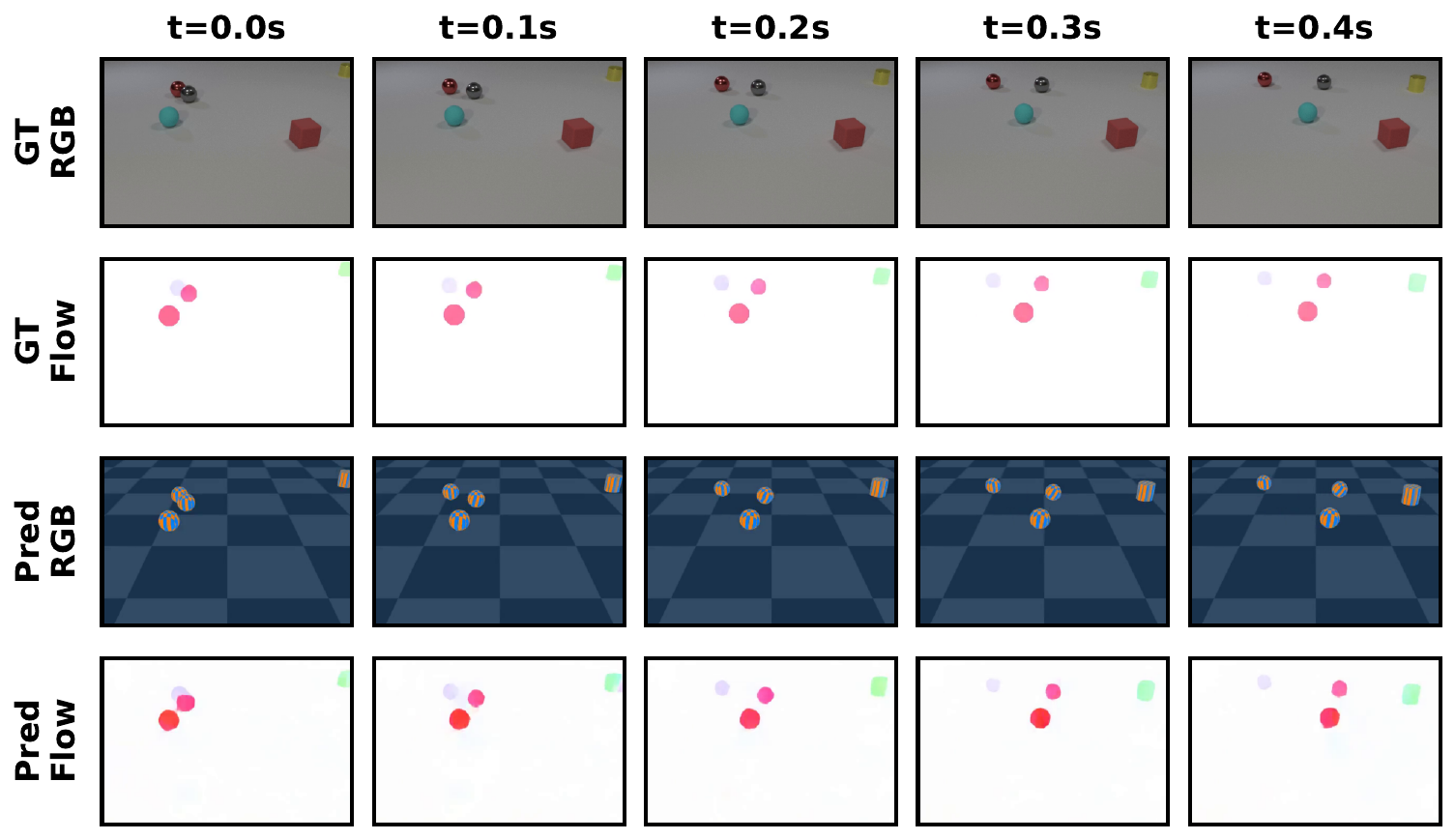}
        \caption{A scene with three rolling balls and one sliding cylinder. The static red box is not reconstructed, as it is not dynamic.}
        \label{fig:eval_real_world_1}
    \end{subfigure}
    % \vspace{-2mm}
    % \hfill
    % \vspace{-2mm}
    % \begin{subfigure}[b]{0.43\textwidth}
    %     \centering
    %     \includegraphics[width=\linewidth]{figures/clevrer_T5/video_15810_Scale3_AS_Reason_IoU_53.pdf}
    %     % \caption{}
    %     \label{fig:eval_real_world_3}
    % \end{subfigure}
    % \hfill
    % \begin{subfigure}[b]{0.48\textwidth}
    %     \centering
    %     \includegraphics[width=\linewidth]{figures/clevrer_T5/video_15890_Scale2_ID_Reason_IoU_22.pdf}
    %     \caption{}
    %     \label{fig:eval_real_world_4}
    % \end{subfigure}
    \vspace{-2mm}
    \caption{\textbf{Zero-shot generalization between engines, from MuJoCo to Blender.} We train \ours~on MuJoCo data and evaluate it on CLEVRER~\citep{yi2019clevrer}. For each example, we show (from top to bottom) (1) the original RGB video, (2) the ground truth optical flow, (3) our model's reconstructed video, and (4) the optical flow of our reconstruction. }
    \vspace{-2mm}
    \label{fig:eval_clvrer}
\end{figure}

\vspace{0.5em}
\noindent\textbf{Comparison with Baselines}
In Table~\ref{tab:eval_clevrer_baseline},~\ourss consistently demonstrates superior performance compared to baseline models. Furthermore, the previous observations on our synthetic dataset hold true: (i) the model using optical flow inputs outperforms that using RGB videos, and (ii) incorporating reasoning further increases motion reconstruction accuracy, e.g., full-sequence segmentation map IoU increases from $0.24$ to $0.29$, a $21\%$ relative improvement. 
Complementary to numerical results, Figure~\ref{fig:eval_clvrer} shows that~\ourss accurately captures multi-object motion trajectories, even in an unseen domain such as CLEVRER.

% Despite strong performance in greedy decoding, out-of-domain evaluation reveals a fundamental misalignment between the training objective and the ultimate goal of physical simulation. To address this, we reframe the task from ``predicting the single most probable sequence" to ``finding the best sequence from a diverse set of candidates." 

\vspace{0.5em}
\noindent\textbf{Test-Time Enhancement.}
% At test time, we perform sampling-based search over the model’s learned parameter space.
% This reframes the objective from token-level likelihood to instance-level physical accuracy, allowing the model to discover parameter configurations that better reproduce the overall observed dynamics.
% We adopt a best-of-$N$ evaluation scheme: for each test case, we generate $N=32$ diverse predictions with temperature $0.1$ and report best@$k$ performance.
% This metric reflects the model’s ability to recover high-quality physical reconstructions when given multiple attempts.
We evaluate the model's performance with different testing time strategies. As shown in Table~\ref{tab:eval_CLEVRER_sampling_rl_simplifed}, the vanilla \ours~model gains a marginal improvement by sampling more scene configurations. For example, the full-sequence IoU increases from 0.24 (with the greedy sampling approach) to 0.28 (the best out of  32 sampled configurations), a 14\% increase.

Consistent with earlier findings, the motion-reasoning variant significantly outperforms the base model, and using best-of-32 sampling further boosts performance: the first-frame IoU increases from $0.30$ to $0.38$ (+27\%), and the full-sequence optical flow EPE decreases by 13\%. 
These relative gains are larger than those achieved by the vanilla model under the same sampling strategy.
We hypothesize that the intermediate motion-reasoning step provides a more structured and physically meaningful representation, which enables sampling to explore a broader and effective set of plausible solutions rather than drifting into implausible regions of the parameter space. 
This broader yet more guided search helps resolve long-tailed errors and yields higher-quality reconstructions.

Lastly, we perform evolutionary search with an initialization from the best-of-32 sample. This method yields the highest accuracy for the full sequence, partly thanks to the quality initialization. This result shows that CMA-ES is the method of choice for optimal accuracy during test-time. 
\begin{table}[thbp]
\centering
\scriptsize
\caption{ \textbf{Performance of real-world rigid-body motion reconstruction.} We evaluate \ours~on real-world video dataset. \ours~successfully generalizes from synthetic training to real scenes.
Incorporating motion reasoning improves segmentation and flow alignment, while Best-of-32 sampling further refines accuracy.
CMA-ES optimization provides the best full sequence alignment results.}
\vspace{-2mm}
\begin{tabular}{l cc|cc}
\toprule
 & \multicolumn{2}{c|}{\textbf{Segmentation Map IoU (↑)}} & \multicolumn{2}{c}{\textbf{Optical Flow EPE (↓)}} \\
 & First Frame & Full Seq. & First Frame & Full Seq. \\
\midrule
\ours 
& 0.57 & 0.26 & 1.62 & 0.67 \\
+ Motion Reasoning  & 0.54 & 0.29 & 1.39 & 0.58 \\
+ + Best@32 
& \textbf{0.72} & \underline{0.41} & \textbf{1.06} & \underline{0.46} \\
+ + CMA-ES 
& \underline{0.57} & \textbf{0.65} 
& \underline{1.26} & \textbf{0.36} \\
\bottomrule
\end{tabular}
\label{tab:eval_real_world_eval}
\end{table}

\begin{figure}[th]
    % \centering
    % \begin{subfigure}[b]{0.45\textwidth}
    %     \centering
    %     \includegraphics[width=\linewidth]{figures/real_world_T5/IMG_1956_Scale4_late_AS_Reason_IoU_50.pdf}
    %     \caption{}
    %     \label{fig:eval_real_world_1}
    % \end{subfigure}
    % \hfill
    \begin{subfigure}[b]{0.45\textwidth}
        \centering
        \includegraphics[width=\linewidth]{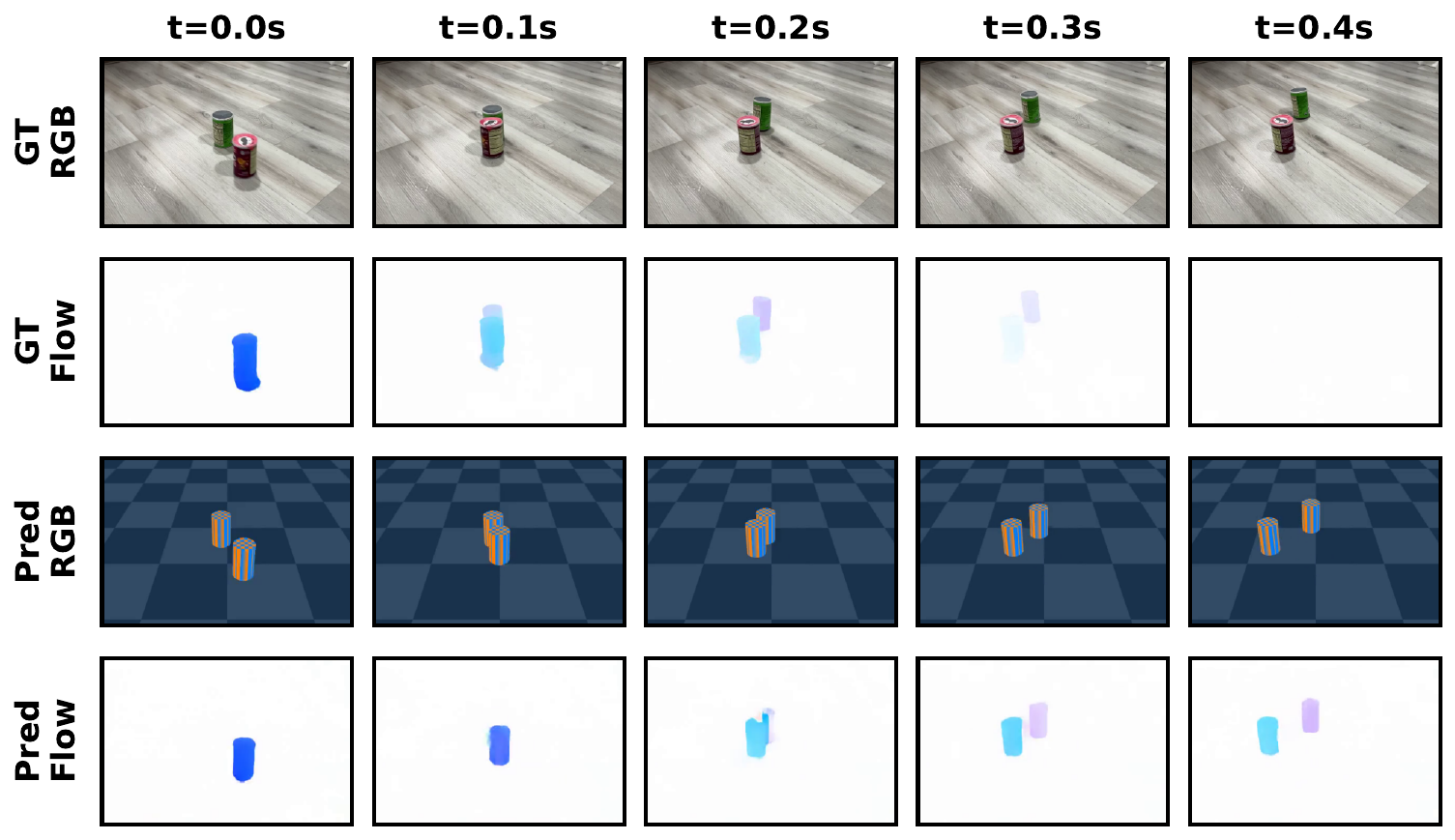}
        \caption{A moving container collides with a stationary one.}
    \end{subfigure}
    % \vspace{-1mm}
    % \hfill
    \begin{subfigure}[b]{0.45\textwidth}
        \centering
        \includegraphics[width=\linewidth]{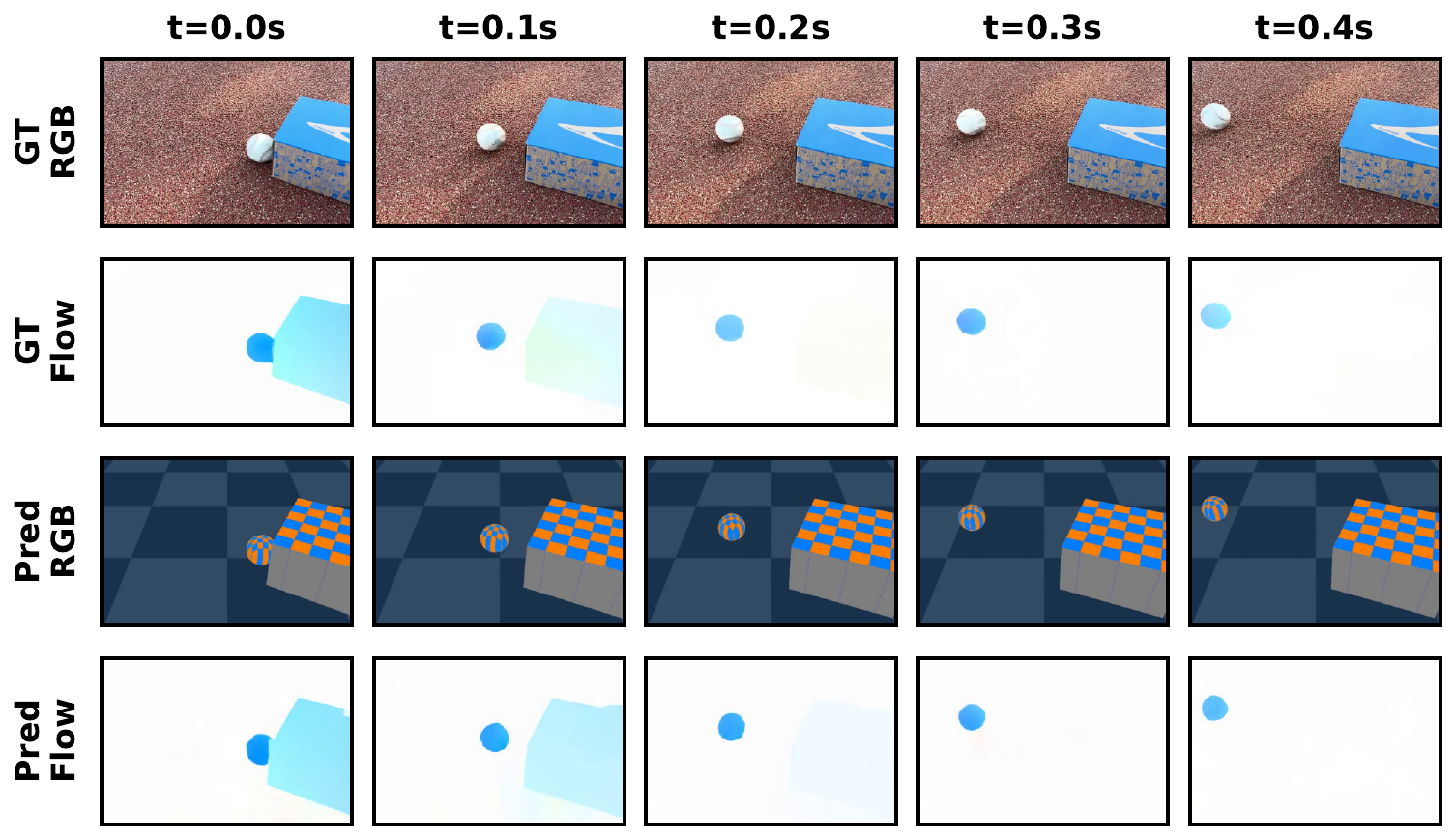}
        \caption{A sliding shoebox collides with a baseball on a running track.}
    \end{subfigure}
    % \hfill
    % \begin{subfigure}[b]{0.45\textwidth}
    %     \centering
    %     \includegraphics[width=\linewidth]{figures/real_world_T5/IMG_2191_Scale4_late_ID_NoReason_IoU_66.pdf}
    %     \caption{}
    % \end{subfigure}
    % \begin{subfigure}[b]{0.45\textwidth}
    %     \centering
    %     \includegraphics[width=\linewidth]{figures/real_world_T5/IMG_2244_Scale4_late_AS_NoReason_IoU_20.pdf}
    %     \caption{}
    % \end{subfigure}
    % \hfill
    % \begin{subfigure}[b]{0.45\textwidth}
    %     \centering
    %     \includegraphics[width=\linewidth]{figures/real_world_T5/IMG_2246_Scale4_ID_Reason_IoU_26.pdf}
    %     \caption{Two shoeboxes collide on a  basketball court.}
    % \end{subfigure}    
    \vspace{-1mm}
    \caption{\textbf{Motion capture for real-world videos.} \ours~is able to reproduce motion trajectory and object location on real-world surfaces and complex lighting. It can also capture multi-body collision dynamics despite the domain gap between synthetic and real data.}
    \label{fig:eval_real_world}
\end{figure}

\subsection{Real-World Applications}

In this section, we evaluate \ours in the real world. Additional results on physically plausible video editing are provided in Section~\ref{app:editing}.

\vspace{0.5em}
\noindent\textbf{Dataset.}
We collected real-world videos using an iPhone~13 and Canon cameras in landscape orientation.
To assess robustness to diverse surface conditions, we collected data for multiple environments, including indoor floors, outdoor running tracks, and basketball courts.
Test objects include everyday items such as shoe boxes, balls, massage roll, cookie containers, and some irregular-shaped objects such as apples. The dataset details are provided in Appendix~\ref{appendix:real_world_stats}.

\vspace{0.5em}
\noindent\textbf{Results.}
As shown in Table~\ref{tab:eval_real_world_eval}, the motion reasoning variant improves segmentation IoU by 12\% and flow EPE by 13\%, while best-of-32 sampling further enhances motion accuracy. Evolutionary search provides the largest gains in the metrics. 
Qualitatively, Figure~\ref{fig:eval_real_world} shows that our model captures the trajectories and locations of two-object motion precisely, implying a high level of accuracy in the estimated initial states and physics parameters. We provide more real-world examples and failure analysis in Appendix~\ref{appendix:real_world_result}.
\section{Conclusion}
We have presented a novel viewpoint on the problem of predicting physics configurations for rigid-body motion from monocular videos. Our main contribution is a general structured textual representation of the physics states and parameters for a wide range of motion dynamics and object interaction. In addition, we trained \ours, a vision–language model to generate physics configurations and camera geometry in a structured textual format.
We also incorporate motion reasoning and test-time optimization techniques to enhance our model's accuracy. Being trained on $400 K$ synthetically generated scenes in MuJoCo, our model shows robust generalization across rendering engines and to real-world data, and consistently outperforms off-the-shelf vision–language models. Our results demonstrate a promising line of research on using language modeling to provide a common physics representation for physics perception and physics simulation.

% Instead of predicting a fix-length parameter vector, we formulate physics inference as a conditional text generation problem, where the model produces structured textual configurations encoding object geometries, material and inertial properties, initial states, and camera poses—directly executable by a physics engine.
% Training on 400 K procedurally generated MuJoCo scenes, our model achieves robust performance across object counts and motion regimes while remaining invariant to scene semantics. Incorporating structured motion reasoning and test-time optimization, including sampling and evolutionary search, further improves physical fidelity and generalization.
% \ourss surpasses state-of-the-art vision–language baselines by over $7\times$ in segmentation IoU on CLEVRER.
% We also curate and release a real-world benchmark of 104 rigid-body motion videos, annotated with optical flows and segmentation masks, to advance research in physics-aware video understanding.
% On this dataset, \ourss transfers effectively to complex real-world scenes with irregular shapes and unconstrained viewpoints.
% Our results highlight the promise of language-driven physics understanding as a unifying pathway between perception and simulation.

{
    \small
    \bibliographystyle{ieeenat_fullname}
    \bibliography{main}
}

\appendix
\clearpage
\newpage
\setcounter{page}{1}
\maketitlesupplementary
\section{Method Details}

\subsection{YAML-to-XML Conversion and Initialization}
\label{app:yaml_xml}

Our dataset uses YAML as the canonical representation of the full scene configuration. 
The YAML file contains both static attributes (e.g., object geometries, masses, friction coefficients, camera pose) and dynamic initial states (linear and angular velocities). 
MuJoCo, however, accepts model definitions only in its XML-based scene description format, which encodes static model properties but does not support specifying initial velocities.

To run each simulation, we therefore proceed in two steps. 
First, we convert the static components of the YAML configuration into a MuJoCo XML file, defining the bodies, inertial properties, joints, geoms, and camera. 
Second, we load this XML into MuJoCo, initialize the engine and then set all remaining dynamic quantities (e.g., initial linear and angular velocities) directly in the simulator state before simulation rollout. 

\subsection{Dataset Preparation}
\label{app:dataset_preparation}

% \noindent \textbf{Synthetic Dataset Curation}
% To ensure physically plausible and visually interesting interactions, we apply several filters that discard (i) scenes with overlapping objects at initialization using MuJoCo's collision detection, (ii) scenes where objects remain outside the camera's field of view throughout the sequence; and (3) scenes with objects that are too small to be visually distinct (total pixel area $< 8000$ pixels). Each simulation is rendered for 1~second at 30~FPS with eight physics substeps per frame, producing images of $480{\times}320$ resolution.  

\noindent \textbf{MuJoCo Rendering Details.}  
Each simulation is rendered for one second at 30~FPS, with eight physics steps per frame. The output resolution is $480 \times 320$ pixels.  
We introduce randomized lighting conditions with varying shadow configurations and apply diverse object textures, including checkerboard and gradient patterns, to improve robustness to visual appearance.  
During simulation, we record RGB frames, segmentation masks, contact information, and full state histories, which are later used to generate structured visual reasoning annotations.

Optical flow estimation is often unreliable on smooth or textureless surfaces. To mitigate this, we fill the background and floor with natural scene images, which introduce sufficient texture and yield substantially higher-quality flow fields. Optical flow is computed between consecutive frames (30 FPS), producing 29 flow maps per video; we then sample every third frame to obtain the inputs used for model training.

\vspace{0.4em}
\noindent \textbf{Structured Visual Reasoning.}  
To ensure consistent and interpretable spatial reasoning, we discretize continuous object positions and motions into categorical linguistic descriptors along three spatial axes.  
For example, positions along the $x$-axis are quantized into seven categories: far left ($x < -2$), moderately left ($-2 \leq x < -1$), slightly left ($-1 \leq x < -0.5$), near center ($-0.5 \leq x < 0.5$), slightly right ($0.5 \leq x < 1$), moderately right ($1 \leq x < 2$), and far right ($x \geq 2$).

Segmentation maps are used to determine whether each object is initially visible and to record when it leaves the camera’s field of view.  
During simulation, we log collision and contact histories from the physics engine to capture fine-grained interaction events such as ground contact and inter-object collisions.  
We also analyze object state trajectories to identify when each object comes to rest, enabling precise annotation of temporal dynamics such as motion duration and stopping time.
To increase linguistic diversity, the transformation script converts these auxiliary signals into textual descriptions using multiple paraphrased sentence templates while preserving structural consistency. 

\vspace{0.4em}
\noindent \textbf{Example of Structured Visual Reasoning.} \label{appendix:reasoning}
An example of synthetic data is shown in Table~\ref{tab:reasoning_appendix_example}. The model analyzes object properties, motion patterns, and physical interactions in natural language, which serves as an intermediate step to improve parameter estimation accuracy.

\subsection{Details on In-Context Example Preparation}\label{app:icl_examples}

To guide the models effectively, we construct three in-context examples that (i) include all primitive shapes present in our dataset and (ii) cover the full range of physical parameters by selecting configurations at the minimum and maximum values of the training distribution. Each example consists of ten video raw RGB frames paired with its YAML-based scene configuration.

For CLEVRER~\cite{yi2019clevrer}, we first select three target example videos and manually annotate three corresponding scene configurations. The examples are chosen to (i) include all three primitive shapes and (ii) span the minimum and maximum values of our physical-parameter ranges, ensuring that the model does not extrapolate beyond the provided examples. To maintain fidelity, we iteratively refine each configuration annotation so that the resulting simulated motions and object trajectories closely match those in the target videos.

\subsection{Details on Preference Optimization} \label{app:method_preference_optimization}

\vspace{0.4em}
\noindent\textbf{Preliminary: Preference Rank Optimization.}
Following the Bradley–Terry formulation~\citep{bradley1952rank}, a reward model (RM) estimates pairwise preferences by contrasting two responses $y^1$ and $y^2$ for a given input $x$.  
Preference Ranking Optimization (PRO)~\citep{song2024preference} extends this idea by directly fine-tuning the policy $\pi_\theta$, treating it as both the RM and the policy network. The PRO loss is defined as:
\begin{equation}
\mathcal{L}_{\mathrm{PRO}}
= -\log
\frac{e^{r_\pi(x, y^1)}}
{e^{r_\pi(x, y^1)} + e^{r_\pi(x, y^2)}},
\label{eq:pro}
\end{equation}
where $r_\pi(x, y)$ denotes the implicit reward $r_{\pi_{\mathrm{PRO}}}$ for a given input $x$ and candidate response $y^k$. It is defined as the average token-level log-likelihood:
\begin{equation}
r_{\pi_{\mathrm{PRO}}}(x, y^k)
=
\frac{1}{|y^k|}
\sum_{t=1}^{|y^k|}
\log P_\pi\!\left(y^k_t \mid x, y^k_{<t}\right).
\label{eq:rpi_pro}
\end{equation}
Intuitively, $r_{\pi_{\mathrm{PRO}}}(x, y^k)$ measures the normalized sequence log-likelihood (i.e., the mean per-token log-probability) and serves as a scalar proxy for how confidently the model assigns probability mass to the response $y^k$ given $x$.

\vspace{0.4em}
\noindent\textbf{Soft Preference Weighting.}
However, Eq.~\ref{eq:pro} assumes a \emph{one-hot} preference—one response is strictly preferred ($y^1\!\succ\!y^2$) while the other is not.  
In our setting, this binary assumption is overly rigid: two simulated rollouts may each excel in distinct aspects (e.g., one accurately reproduces geometry while the other better matches damping or velocity).  
To capture such nuanced trade-offs, we introduce a \emph{soft preference weighting} that transforms the simulator-derived rewards into continuous targets:
\[
\tilde{r}(y^i)
=
\frac{e^{s(y^i)/\tau}}
{\sum_{j \in \{1,2\}} e^{s(y^j)/\tau}},
\]
where $\tau$ is a temperature and $s(\cdot)$ denotes the simulation-derived score function, i.e., segmentation IoU. Thus, the optimization objective then becomes:
\begin{equation}
\label{eq:softpro}
\begin{split}
\mathcal{L}_{\mathrm{soft\text{-}PRO}}
= -\!\!\sum_{i \in \{1,2\}} \tilde{r}(y^i)
\log
\frac{e^{r_\pi(x, y^i)}}{
e^{r_\pi(x, y^1)} + e^{r_\pi(x, y^2)}}
\end{split}
\end{equation}
which can be interpreted as a \emph{reward-weighted cross-entropy}—analogous to replacing a binary BCE loss with a soft-label CE loss.  
This formulation better accommodates partially correct rollouts and encourages the policy to allocate probability mass in proportion to their normalized simulator rewards, leading to smoother and more stable test-time adaptation. This approach bears similarity with soft-preference concept in previous work~\cite{furuta2024geometric, sharifnassab2024soft}

\section{More Qualitative Results}

For direct comparison of physical parameters, we show initial state error on synthetic eval set in Table~\ref{tab:initial_states}. 

\begin{table}[h]
\centering
\caption{Initial States Estimation (MAE $\downarrow$).}
\vspace{-3mm}
\label{tab:initial_states}
\footnotesize
\begin{tabular}{lcc}
\toprule & Position & Velocity \\
\midrule
InternVL3 / Qwen2.5 / Claude4 & 3.05 / 3.39 / 2.55 & 4.38 / 3.32 / 3.85 \\
Ours / Ours+Analysis & 2.20 / \textbf{2.16} & 3.14 / \textbf{2.94} \\
\bottomrule
\end{tabular}
\end{table}

\begin{table*}[thbp]
\captionof{table}{
\textbf{Example Synthetic Training Data Instance.} 
We illustrate the target output format for both the vanilla and reasoning-augmented variants of our model. 
The blue box shows the structured YAML configuration.
% , which specifies object types, geometric dimensions, initial physical states, and global simulation parameters. 
The red box shows the corresponding natural-language reasoning describing object motions and interactions. 
In the vanilla setting, the model is trained to generate only the configuration text wrapped within the \texttt{<answer>} tag. 
In the reasoning-enhanced setting, the model first outputs the reasoning text enclosed by \texttt{<think>} tags, followed by the configuration text within \texttt{<answer>} tags.
}
\centering
\small
\renewcommand{\arraystretch}{1.2}
\setlength{\tabcolsep}{6pt}
\begin{tabular}{p{0.97\linewidth}}
\vspace{-3mm}
\inlineboxBlue{configuration} = 
\begin{tcolorbox}[colback=blue!3!white, colframe=blue!40, left=2mm, right=2mm, top=0.5mm, bottom=0.5mm, boxsep=0mm]
\begin{lstlisting}[style=yaml]
- type: box
  name: box_0
  size: [1.0, 0.5, 0.4]
  state:
    angular_velocity: [0, 0, 0]
    linear_velocity: [4.3, 2.5, 0.0]
    orientation: [0.97, 0.0, 0.0, 0.23]
    position: [-5.0, -0.3, 0.4]
  physics:
    friction: [1.1, 0.3]
    mass: 1.0
    damping: -4
- type: cylinder
  name: cylinder_0
  radius: 0.3
  height: 0.5
  state:
    angular_velocity: [0, 0, 0]
    linear_velocity: [-0.1, -0.5, 0.0]
    orientation: [0.69, 0.69, 0.15, 0.15]
    position: [-0.5, 0.8, 0.3]
  physics:
    friction: [0.5, 0.3]
    mass: 1.0
    damping: -4
- type: camera
  fovy: 45
  orientation: 45
  position: [0, -2, 3.5]
- type: gravity
  gravity: [0, 0, -7.0]
\end{lstlisting}
\end{tcolorbox}

\inlineboxRed{reasoning} = 
\begin{tcolorbox}[colback=red!3!white, colframe=red!40, left=2mm, right=2mm, top=0.5mm, bottom=0.5mm, boxsep=0mm]
\begin{lstlisting}
This physics simulation showcases objects interacting under realistic physics.

- Box_0: A cuboid with dimensions 1.0 x 0.5 x 0.4 m, mass 1.0 kg, positioned leftmost and in the 
foreground, close to the surface. It moves at 4.97 m/s (rightward and forward), stops at t=0.4s, 
and stays grounded from 0.00-0.90s. Collides with cylinder_0 at t=0.1s. 

- Cylinder_0: A solid column (radius=0.3m, height=0.5m, mass=1.0kg) located at the X-axis origin, 
near the camera and base plane. Moves at 0.51 m/s, momentum 2.79 m/s until end, collides with 
box_0 at t=0.1s. 

- Observation Data: Visible entities: box_0, cylinder_0. Both visible in 10/10 frames.

- Dynamic Interactions: Contact event between cylinder_0 and box_0 at t=0.1s.
\end{lstlisting}
\end{tcolorbox}

% {Training Targets for Vanilla \ours}: 
\textbf{Target Sequence (Vanilla \ours):}
\begin{center}
\vspace{-2mm}
\texttt{\textcolor{black}{<answer>}} 
\inlineboxBlue{configuration}
\texttt{\textcolor{black}{</answer>}}
\end{center}

\textbf{Target Sequence (\ourss\textit{+ Motion Reasoning}):}
% {Training Targets for \ourss\textit{+ Reasoning}}:
\begin{center}
\vspace{-2mm}
\texttt{\textcolor{black}{<think>}} 
\inlineboxRed{reasoning}
\texttt{\textcolor{black}{</think>}}
\textbackslash n\textbackslash n
\texttt{\textcolor{black}{<answer>}} 
\inlineboxBlue{configuration}
\texttt{\textcolor{black}{</answer>}}
\end{center}

\end{tabular}
\label{tab:reasoning_appendix_example}
\end{table*}
% \clearpage
\section{Cross-Engine Generalization}

\begin{table*}[t]
\centering
\scriptsize
\caption{\textbf{Generalization Across Simulation Engines.} Evaluating transfer from MuJoCo (training) to Blender (CLEVRER~\cite{yi2019clevrer}) demonstrates that \ours~maintains its performance in a zero-shot setting. Despite domain shifts in rendering and dynamics, incorporating structured motion description consistently improves segmentation IoU and optical flow EPE. 
Note that some off-the-shelf models achieve low optical flow EPE by generating fewer objects than
present in the scene, which artificially reduces motion and lowers EPE compared to models attempting more complete and realistic physics
simulation. 
\textbf{Best} and \underline{runner-up} results are highlighted.
% \textcolor{red}{check Table 8 and Table in the manuscript consistency}
}
\begin{tabular}{l c|cc|cc}
\toprule
 & & \multicolumn{2}{c|}{\textbf{Segmentation Map IoU (↑)}} & \multicolumn{2}{c}{\textbf{Optical Flow EPE (↓)}} \\
 & Input Modality & First Frame & Full Sequence & First Frame & Full Sequence \\
\midrule
\rowcolor{lightblue} \multicolumn{6}{l}{\textit{\textbf{VLM Models}}} \\
InternVL3-8B & RGB & 0.01 & 0.02 & 7.12 & 6.10 \\
Qwen2.5-VL-7B & RGB & 0.01 & 0.01 & 9.22 & 7.41 \\
Claude-4-Sonnet & RGB & 0.03 & 0.04 & 6.34 & \textbf{5.43$^\star$} \\
\midrule
\rowcolor{lightblue} \multicolumn{6}{l}{\textit{\textbf{Ours}}} \\
\ours & RGB & 0.43 & 0.19 & 3.68 & 7.13 \\
\ours & Opt. Flow & 0.63 & 0.24 & 3.51 & 6.85 \\
+ Motion Reasoning & Opt. Flow & \textbf{0.67} & \textbf{0.30} & \textbf{2.79} & 5.94 \\
\bottomrule
\end{tabular}
\label{tab:eval_clevrer_baseline_full}
\end{table*}
\begin{table*}[thbp]
\centering
\scriptsize
\caption{\textbf{Evaluation of Test-Time Sampling and Preference Optimization on CLEVRER.}
We compare the base, reasoning-enhanced, and preference-optimized \ourss under greedy decoding and best-of-$N$ sampling.
For each case, 32 samples are generated with a temperature of $0.1$; \textit{Best@1} denotes the average, while \textit{Best@32} reports the best. \textbf{Best} and \underline{runner-up} results are highlighted.
% \textcolor{red}{check Table 8 and Table in the manuscript consistency}
% \textcolor{red}{make the ++ part clear}
}
\vspace{-1mm}
\begin{tabular}{l l l c|ccc|ccc|ccc}
\toprule
& & & \multicolumn{4}{c|}{\textbf{Segmentation Map IoU ($\uparrow$)}} & \multicolumn{6}{c}{\textbf{Optical Flow EPE ($\downarrow$)}} \\
\cmidrule{2-13}
& \multicolumn{3}{c}{First Frame} & \multicolumn{3}{c|}{Full Sequence} & \multicolumn{3}{c}{First Frame} & \multicolumn{3}{c}{Full Sequence} \\
& Greedy & Best@1 & Best@32 & Greedy & Best@1 & Best@32 & Greedy & Best@1 & Best@32 & Greedy & Best@1 & Best@32 \\
\midrule
\multirow{1}{*}{\ours} & 0.63 & 0.63 & 0.67 & 0.24 & 0.24 & 0.28 & 3.66 & 3.65 & 2.92 & 6.91 & 6.86 & 6.21 \\
% \midrule
% \cmidrule{2-13}
\multirow{1}{*}{ + Motion Reasoning (MR)} 
& {0.67} & {0.68} & {0.76} 
& {0.30} & {0.30} & {0.38} 
& {2.92} & {2.93} & {2.22} 
& {5.94} & {5.95} & {5.17} \\
% \midrule
% \cmidrule{2-13}
\multirow{1}{*}{ + MR + PRO}
& {0.68} & \underline{0.69} & \textbf{0.77} 
& {0.31} & {0.31} & \underline{0.39} 
& {2.90} & {2.94} & \underline{1.85} 
& {5.78} & {5.81} & \underline{4.78} \\
\multirow{1}{*}{ + MR + CMA-ES}
& {0.62} & - & - 
& \textbf{0.66} & - & - 
& \textbf{0.13} & - & - 
& \textbf{0.11} & - & - \\
\bottomrule
\end{tabular}
\label{tab:eval_CLEVRER_sampling_rl_simplifed_with_pro}
\end{table*}

\subsection{Dataset Preparation}

We subsample $400$ one-second clips from the validation split of CLEVRER~\citep{yi2019clevrer}.  
Object segmentation masks are obtained from the official release.  
We first clean the masks by checking whether each object’s segmentation changes within the subsampled clip—objects with static masks are treated as non-moving and removed from motion evaluation.  
This information is then used to refine the corresponding optical flow maps, ensuring that static regions are not misinterpreted as motion.

% For a fair comparison with VLM baselines, we manually annotate three few-shot examples in CLEVRER that cover all three geometric primitives and ensure that the corresponding synthetic few-shot videos match the target CLEVRER scenes in both object composition and motion type.

\subsection{Baseline Evaluation Outcomes}

\begin{figure*}[thbp]
    \centering
    \begin{minipage}[b]{0.85\textwidth}
        \centering
        \includegraphics[width=\textwidth]{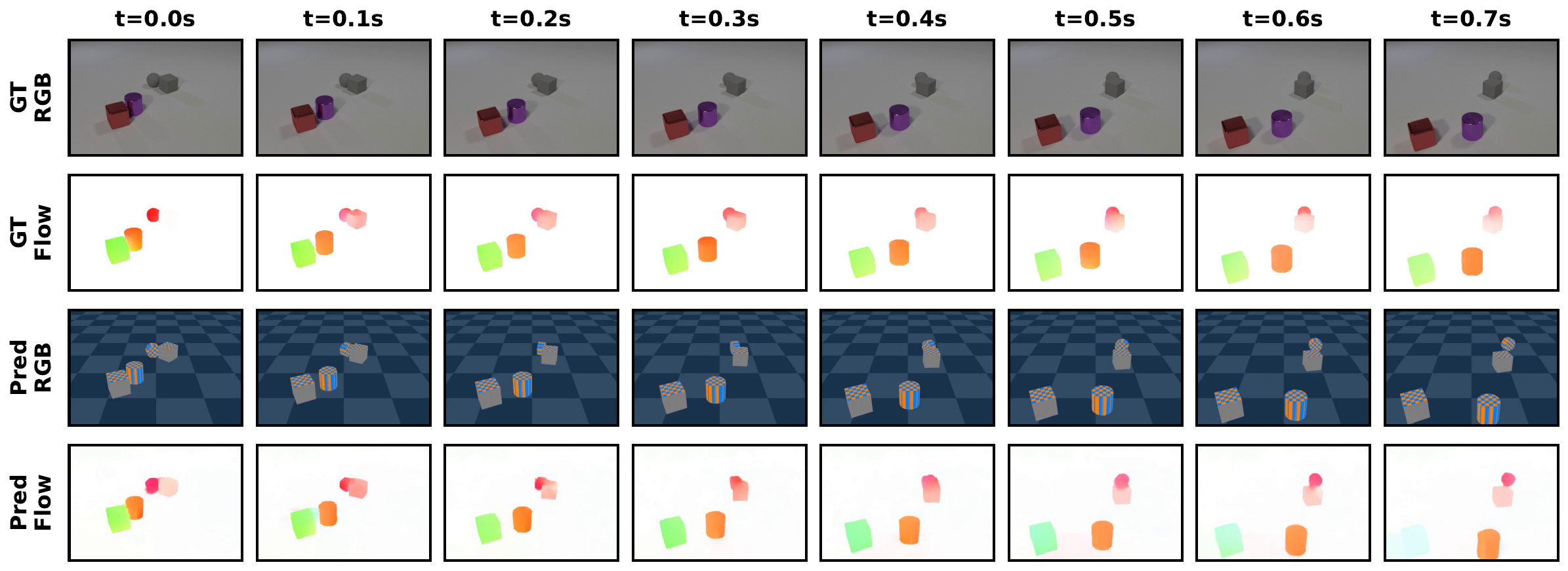}
        % \caption*{(a) Video 15554}
    \end{minipage}
    \hfill
    \begin{minipage}[b]{0.85\textwidth}
        \centering
        \includegraphics[width=\textwidth]{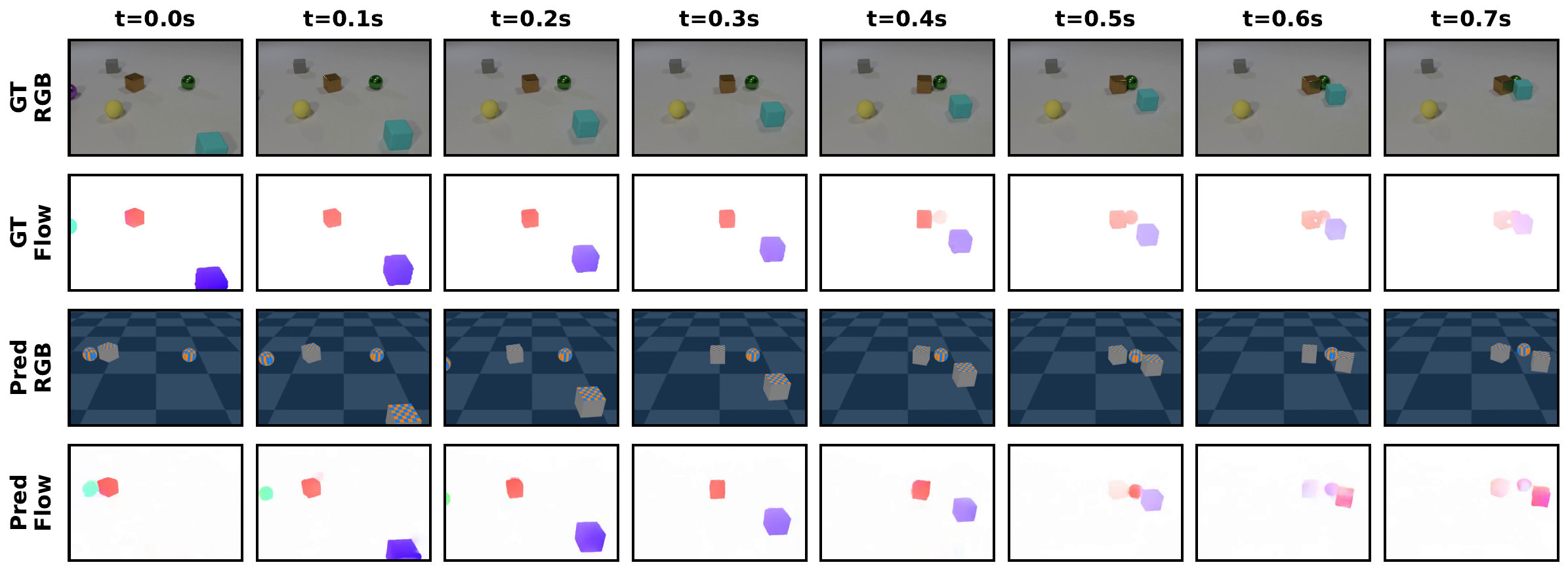}
        % \caption*{(b) Video 15918}
    \end{minipage}
    \hfill
    \begin{minipage}[b]{0.85\textwidth}
        \centering
        \includegraphics[width=\textwidth]{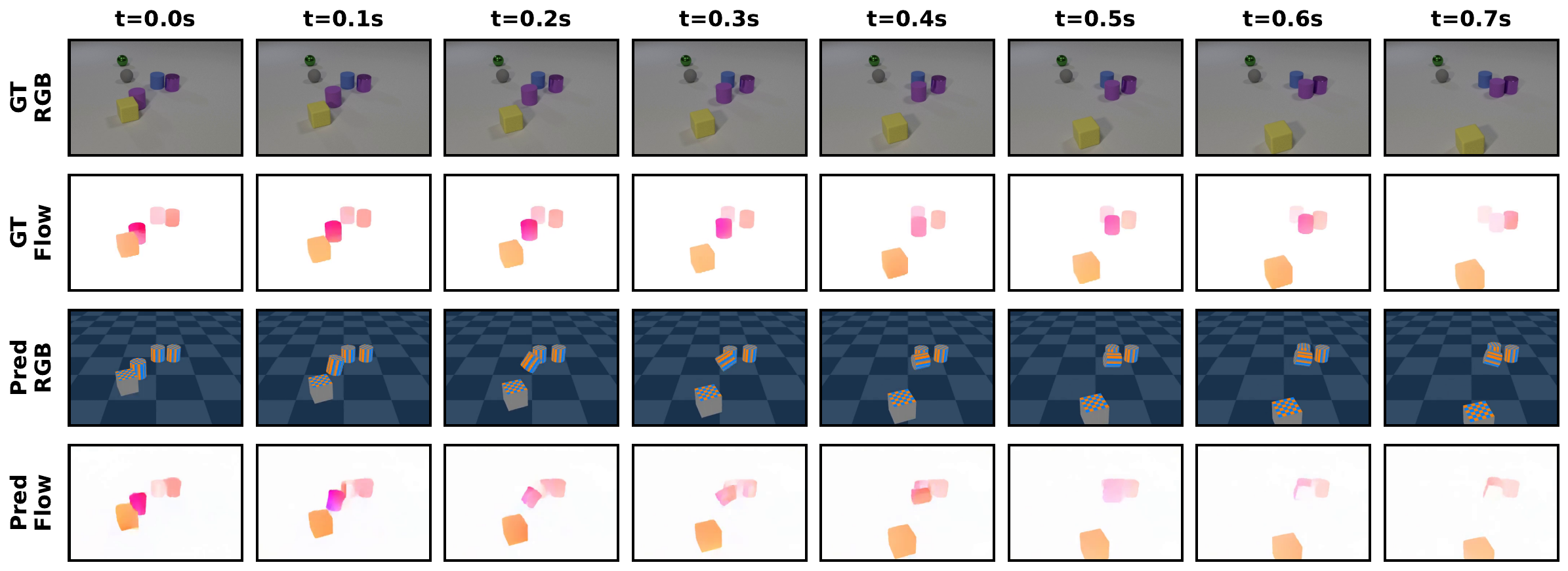}
        % \caption*{(c) Video 15982}
    \end{minipage}
    \hfill
    \begin{minipage}[b]{0.85\textwidth}
        \centering
        \includegraphics[width=\textwidth]{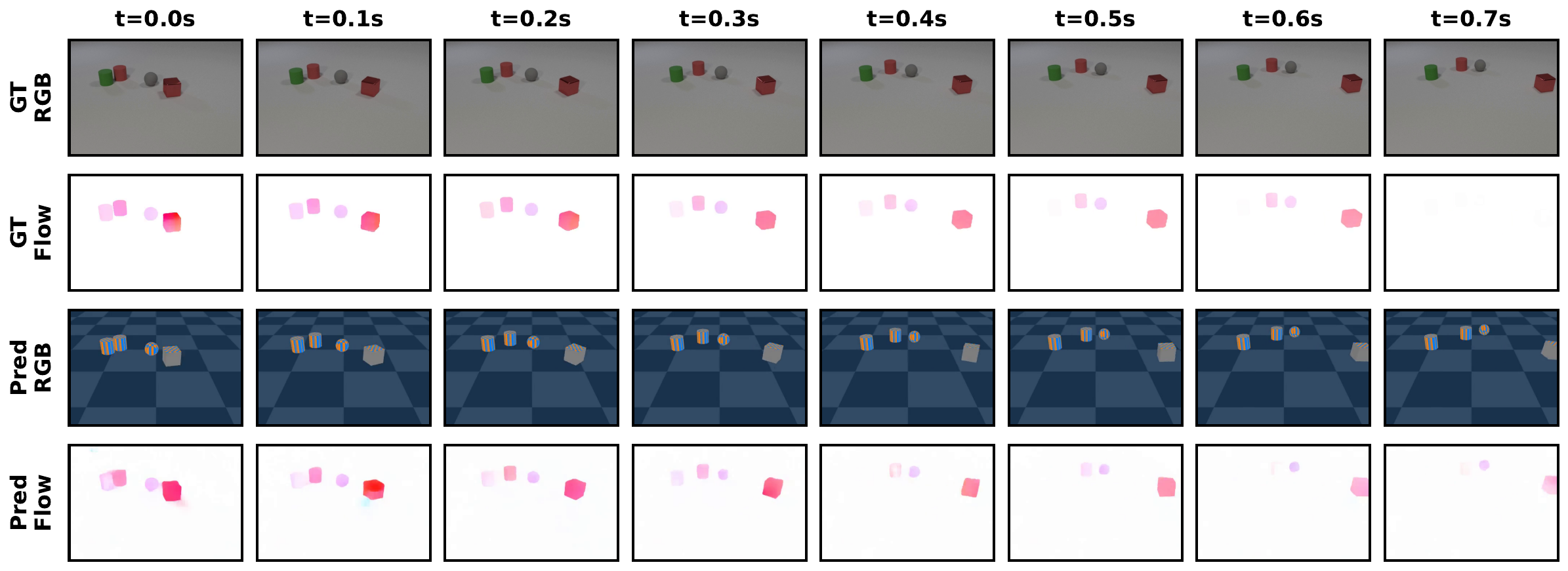}
        % \caption*{(b) Video 15449}
    \end{minipage}
    \caption{CLEVRER Dataset Results.}
    \label{fig:supp_clevrer}
\end{figure*}

The full quantitative evaluations, including optical flows, are shown in Table~\ref{tab:eval_clevrer_baseline_full}. Qualitative results are shown in Figure~\ref{fig:supp_clevrer}. These are the results without using CMA-ES.

\vspace{0.4em}
\noindent\textbf{Test-Time Optimization.}
We therefore explore preference optimization to learn from unlabeled videos as detailed previously in Section~\ref{subsec:method_preference_optimization}. Speficially, we use 1000 cases from CLEVRER training data, generate 32 sampled predictions per case, render rollouts with MuJoCo, and compute segmentation IoU to construct pairwise preferences. We then draw three paired data per case to construct the preference learning dataset and finetune our reasoning model.
As shown in Table~\ref{tab:eval_CLEVRER_sampling_rl_simplifed_with_pro}, preference optimization yields consistent improvements—modest IoU gains (+1\%) and a notable reduction in first-frame EPE (2.22 to 1.85) under the Best@32 metric.
This demonstrates that preference optimization provides a practical route to label-free adaptation, enabling models to refine directly through simulation feedback. However, the marginal gain in performance is less than CMA-ES.

% Second figure with 3 images
% \begin{figure}[H]
%     \centering
%     \begin{minipage}[b]{0.85\textwidth}
%         \centering
%         \includegraphics[width=\textwidth]{figures/s3_supplementary/IoU_17_video_15071_Scale1.pdf}
%         % \caption*{(a) Video 15071}
%     \end{minipage}
%     \hfill
%     \begin{minipage}[b]{0.85\textwidth}
%         \centering
%         \includegraphics[width=\textwidth]{figures/s3_supplementary/IoU_17_video_15449_Scale3.pdf}
%         % \caption*{(b) Video 15449}
%     \end{minipage}
%     \hfill
%     \begin{minipage}[b]{0.85\textwidth}
%         \centering
%         \includegraphics[width=\textwidth]{figures/s3_supplementary/IoU_17_video_15469_Scale3.pdf}
%         % \caption*{(c) Video 15469}
%     \end{minipage}
%     \caption{CLEVRER Dataset Results.}
%     \label{fig:supp_clevrer_2}
% \end{figure}
% \clearpage
\section{Real-World Benchmark and Qualitative Results} 

\subsection{Real-World Dataset Statistics.} \label{appendix:real_world_stats}
Our real-world evaluation set comprises a total of 235 videos capturing diverse rigid-body motion scenarios.
Among them, 155 videos were recorded with an iPhone 13 and 80 videos with a Canon camera.
The iPhone videos were captured at 210 FPS or 240 FPS, while the Canon videos were recorded at 50 FPS.
To create variations in temporal resolution, we downsampled the original recordings by uniformly sampling frames to obtain videos at 25, 30, 50, 60, and 70 FPS.
The resulting frame-rate distribution is: 30 FPS (76 videos), 70 FPS (46), 25 FPS (40), 50 FPS (40), and 60 FPS (33).
Each scene contains between one and five objects, with 139 single-object, 69 two-object, 17 three-object, 6 four-object, and 4 five-object configurations.
The objects span a wide variety of everyday items, including shoebox, drug spray, Pringles can, baseball, tennis ball, tissue box, soccer ball, basketball, pool ball, massage roller, gel container, aerosol can, handcrafted vehicle with wheels, apple, tumbler, soda can, whiteboard eraser, napkin roll, insulation pad, box, banana, and plum.
Among all videos, 86 exhibit object collisions.

We also conduct a pilot human study: two annotators each estimated parameters for three real-world videos with iterative refinement (25 minutes per video). Humans achieved mean IoU of 0.44 and EPE of 1.38, versus \ours’s 0.61 and 0.71, respectively.

\subsection{Qualitative Results.} 
\label{appendix:real_world_result}
\begin{figure*}[thbp]
    \centering
    \begin{subfigure}[b]{0.7\textwidth}
        \centering
        \includegraphics[width=\textwidth]{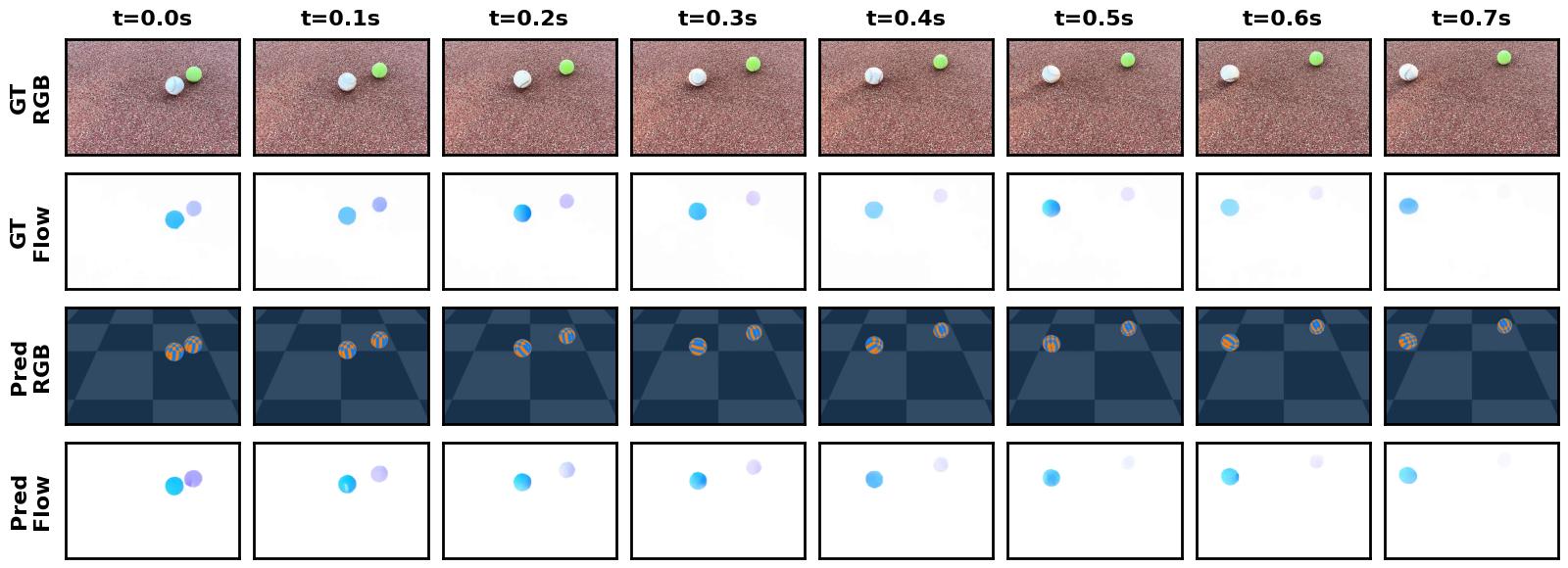}
        \caption{A baseball and a tennis ball moving in different directions.}
    \end{subfigure}
    \hfill
    \begin{subfigure}[b]{0.7\textwidth}
        \centering
        \includegraphics[width=\textwidth]{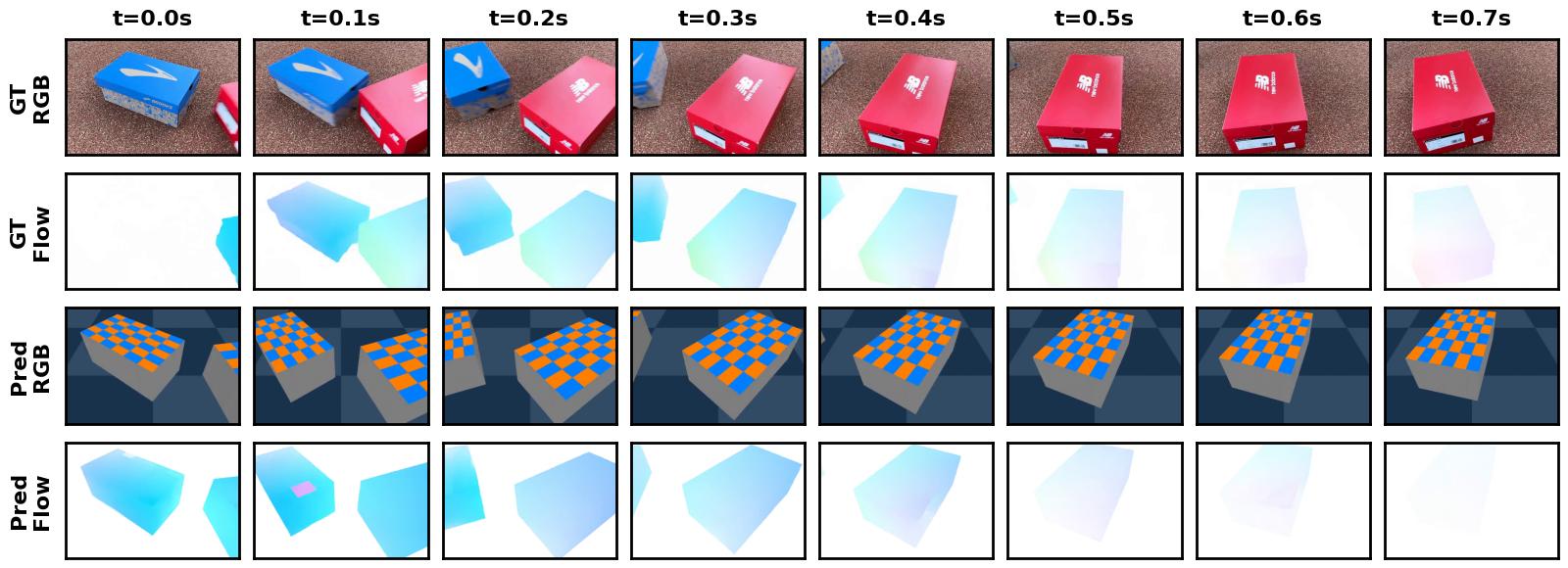}
        \caption{A red shoe box hits a blue shoe box.}
    \end{subfigure}
    \vspace{2mm}
    \begin{subfigure}[b]{0.7\textwidth}
        \centering
        \includegraphics[width=\textwidth]{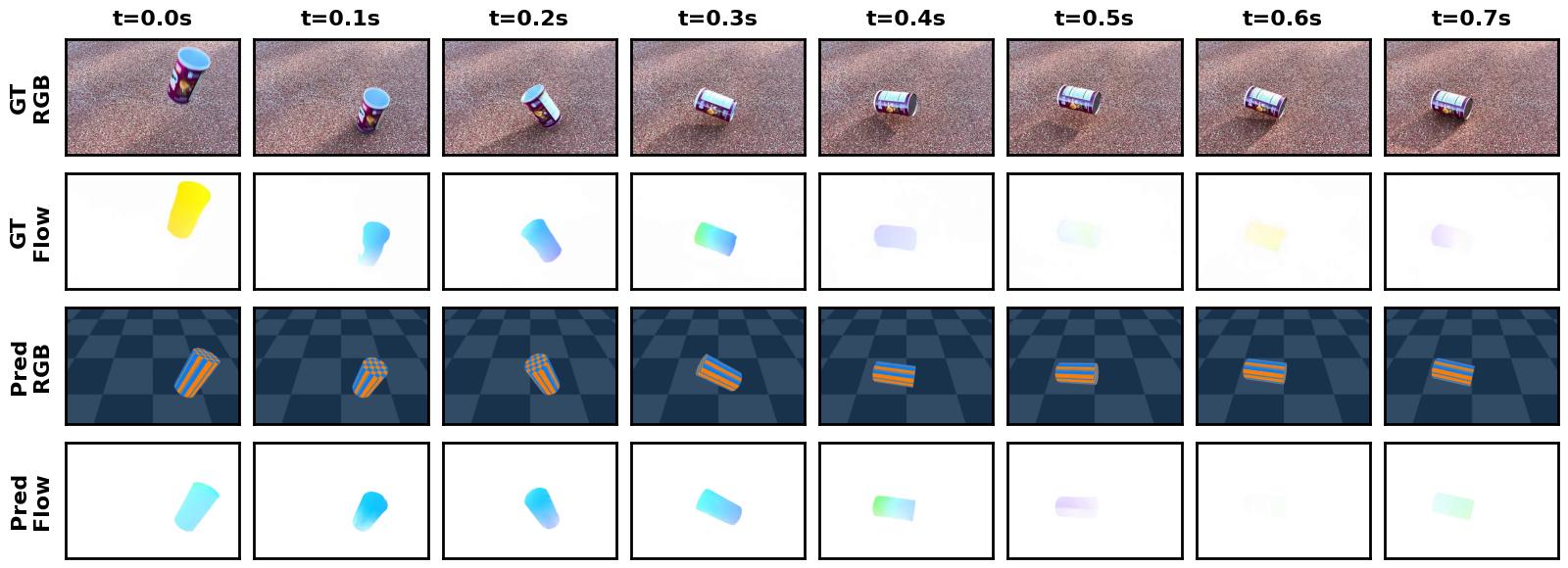}
        \caption{A cylindrical container bouncing on the ground.}
    \end{subfigure}
    \hfill
    \begin{subfigure}[b]{0.7\textwidth}
        \centering
        \includegraphics[width=\textwidth]{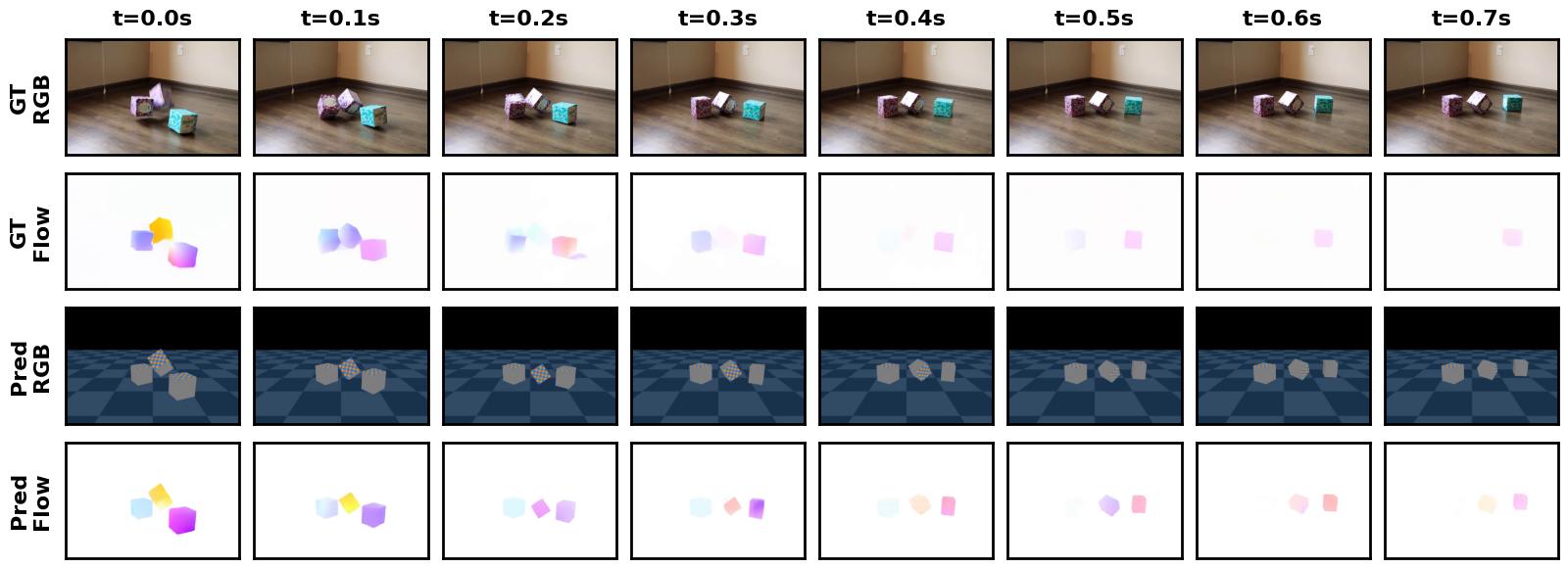}
        \caption{Three tissue boxes dropping on the floor.}
    \end{subfigure}
    \vspace{-1mm}
    \caption{\textbf{Rigid-Body Motion Estimation on Our Real-World Dataset.}
    \ours{} reconstructs physically plausible trajectories from real-world videos of rigid-body motion, capturing object interactions, material properties, and dynamics across diverse conditions.}
    \label{fig:supp_real_world}
    \vspace{-3mm}
\end{figure*}

\begin{figure*}[thbp]
    \centering
    \begin{subfigure}[b]{0.7\textwidth}
        \centering
        \includegraphics[width=\textwidth]{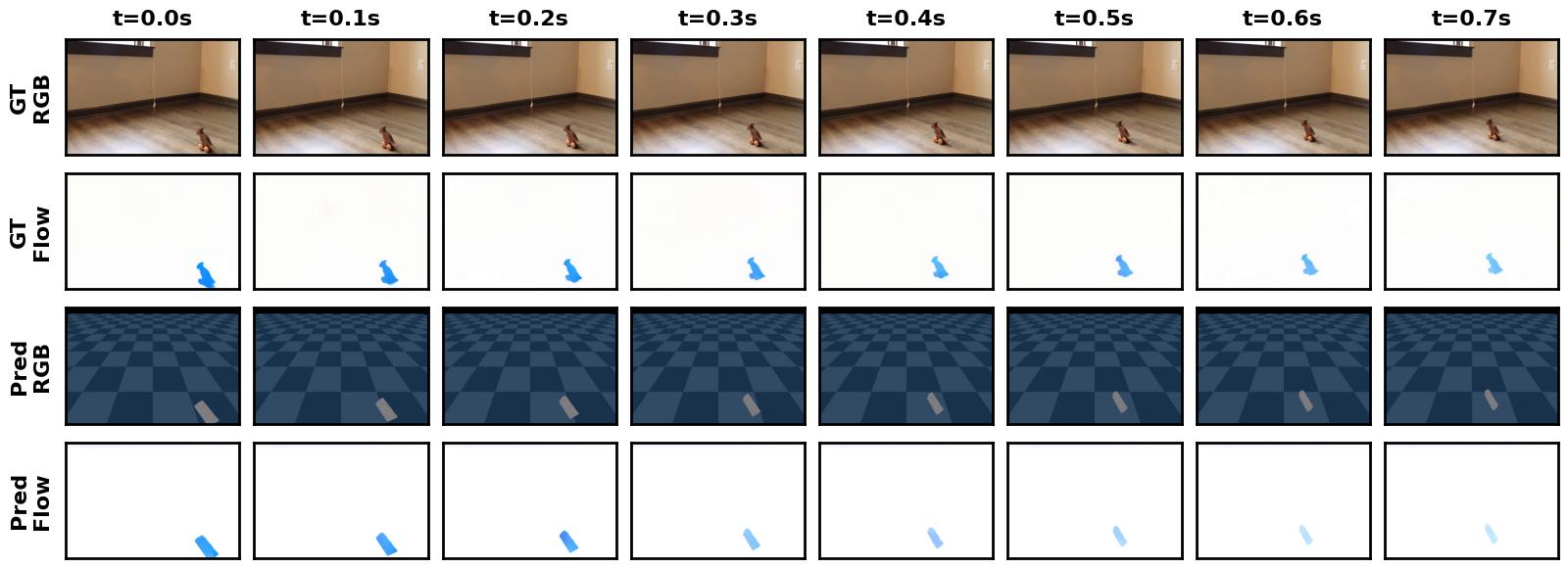}
        \caption{A handcrafted wooden eagle with wheels.}
    \end{subfigure}
    \hfill
    \begin{subfigure}[b]{0.7\textwidth}
        \centering
        \includegraphics[width=\textwidth]{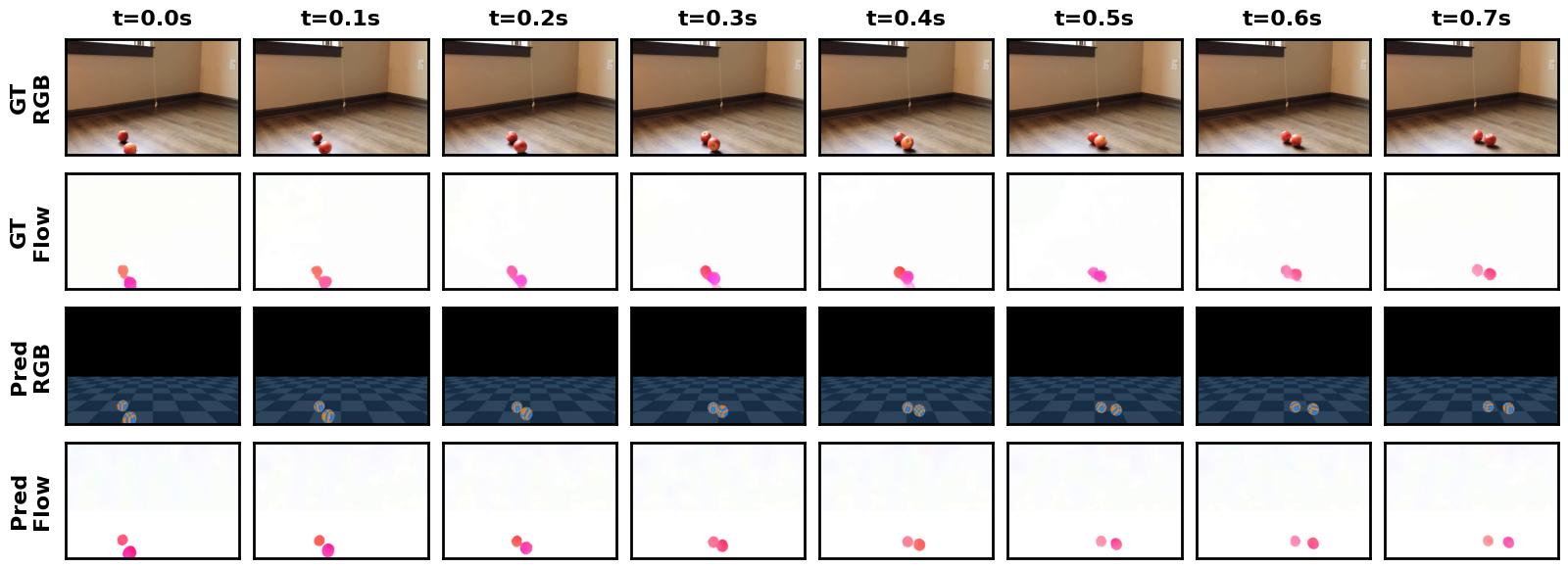}
        \caption{Two moving (rolling) apples collide.}
    \end{subfigure}
    \vspace{2mm}
    \begin{subfigure}[b]{0.7\textwidth}
        \centering
        \includegraphics[width=\textwidth]{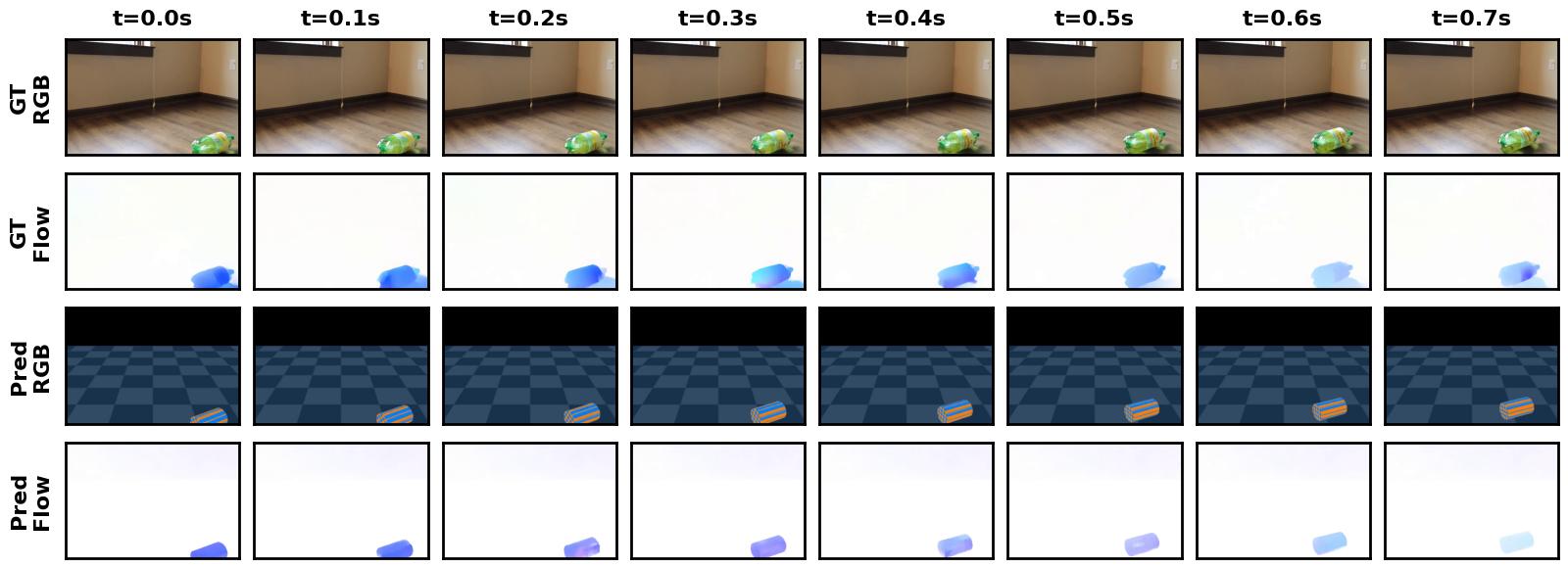}
        \caption{A near-cylindrical soda can with rounded top surfaces rolls slowly .}
    \end{subfigure}
    \vspace{-1mm}
    % \caption{\textbf{Rigid-Body Motion Estimation on Our Real-World Dataset, with a focus of irregular shaped objects}
    \caption{\textbf{Rigid-Body Motion Estimation on Our Real-World Dataset, Focusing on Irregularly Shaped Objects.}}
    % \ours{} reconstructs physically plausible trajectories from real-world videos of rigid-body motion, capturing object interactions, material properties, and dynamics across diverse conditions.}
    \label{fig:supp_real_world_irregular}
    \vspace{-3mm}
\end{figure*}

\begin{figure*}[thbp]
    \centering
    % \begin{subfigure}[b]{0.7\textwidth}
    %     \centering
    %     \includegraphics[width=\textwidth]{figures/real_world_T8_failure/IMG_2193_fps30__CMA.jpg}
    %     \caption{After the collision, the reconstructed red box orientation is different from the ground-truth.}
    % \end{subfigure}
    % \hfill
    \begin{subfigure}[b]{0.7\textwidth}
        \centering
        \includegraphics[width=\textwidth]{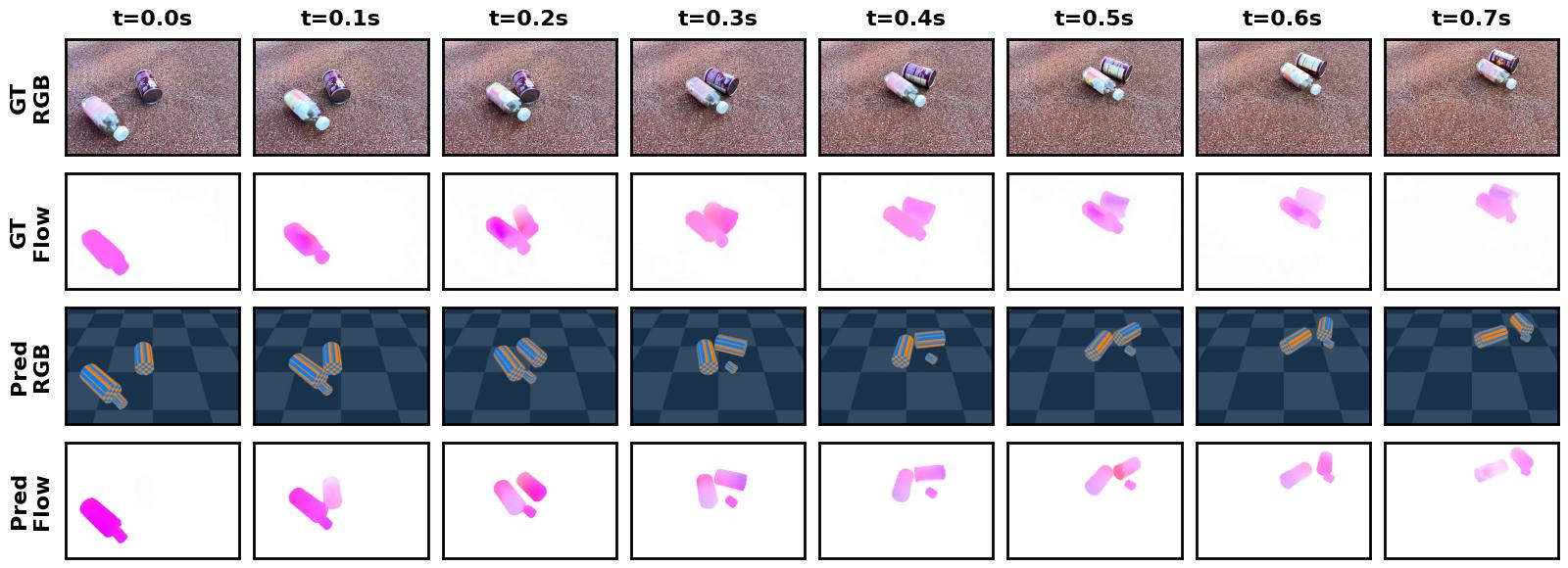}
        \caption{Due to the bottle’s irregular shape, our model approximates it with two cylinders and focuses on faithfully reconstructing the motion in the first frame ($t=0.0$s).}
    \end{subfigure}
    \vspace{2mm}
    \begin{subfigure}[b]{0.7\textwidth}
        \centering
        \includegraphics[width=\textwidth]{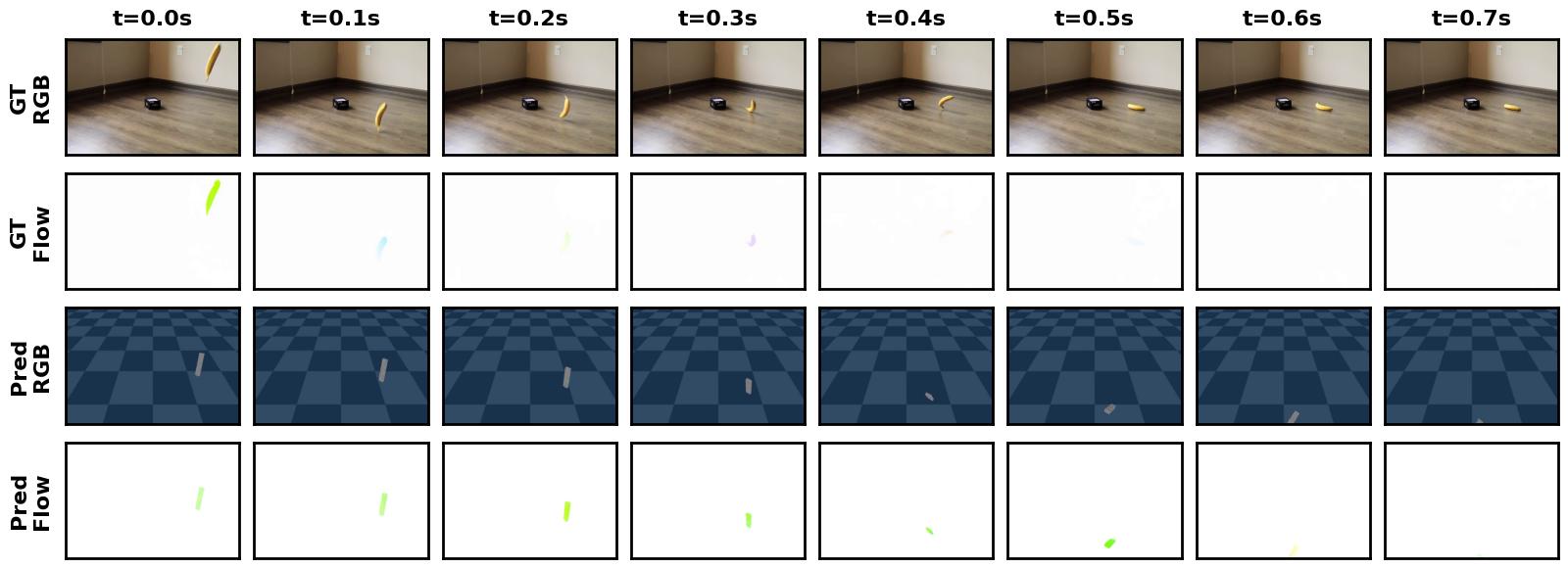}
        \caption{A banana falls to the floor, bounces once, and then comes to rest. Its irregular shape produces an unexpected trajectory, making it difficult to capture object positions and camera poses accurately.}
    \end{subfigure}
    \vspace{2mm}
    \begin{subfigure}[b]{0.7\textwidth}
        \centering
        \includegraphics[width=\textwidth]{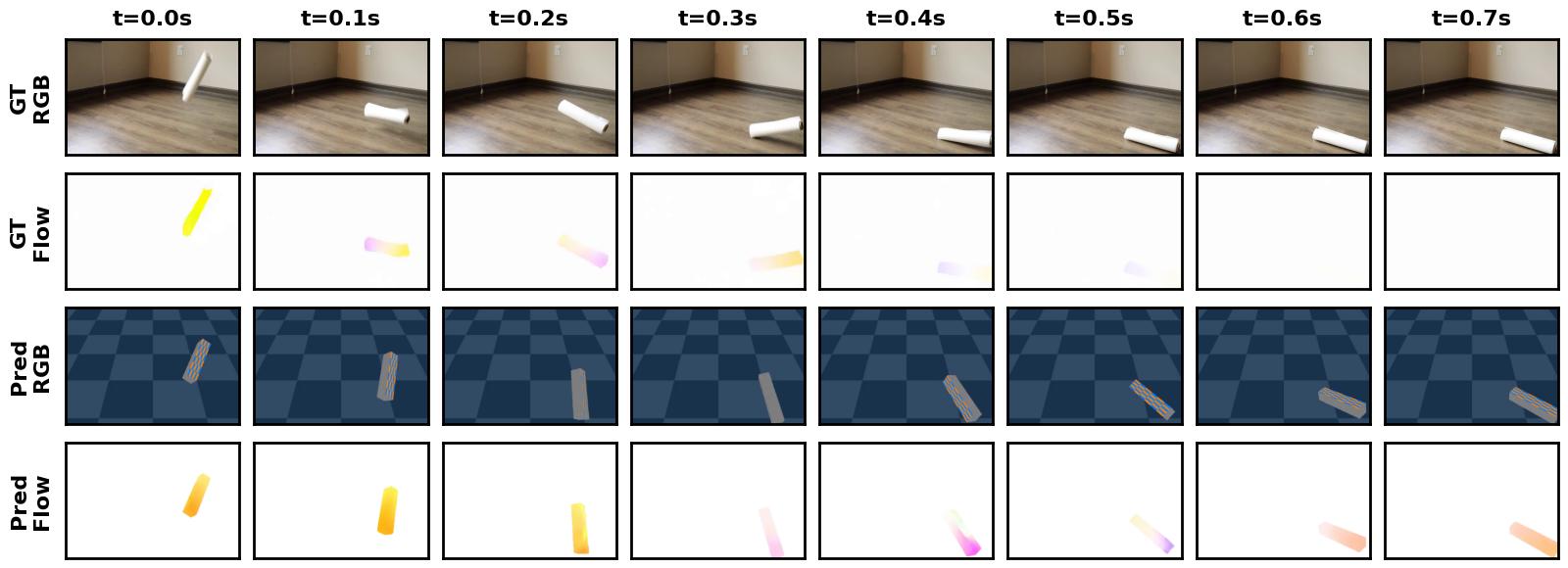}
        \caption{In this case, the model predicts an incorrect primitive shape and an imprecise camera angle.}
    \end{subfigure}
    \vspace{-1mm}
    \caption{\textbf{Rigid-Body Motion Estimation on Our Real-World Dataset, Focusing on Failure Cases.}}
    \label{fig:supp_real_world_failure}
    \vspace{-3mm}
\end{figure*}

In Figure~\ref{fig:supp_real_world} we show more real-world examples; in Figure~\ref{fig:supp_real_world_irregular} we show the results of irregular-shaped objects such as apples, wooden aircraft, and soda cans. We also present failure cases in Figure~\ref{fig:supp_real_world_failure}, where the failure modes comes from irregular shapes that are not seen during training or that results in unexpected bouncing trajectory. In some other cases, the model can predict wrong primitive shapes.

\subsection{Non-object-centric scenes:} 
\begin{figure}[ht]
\centering
\includegraphics[width=0.48\textwidth]{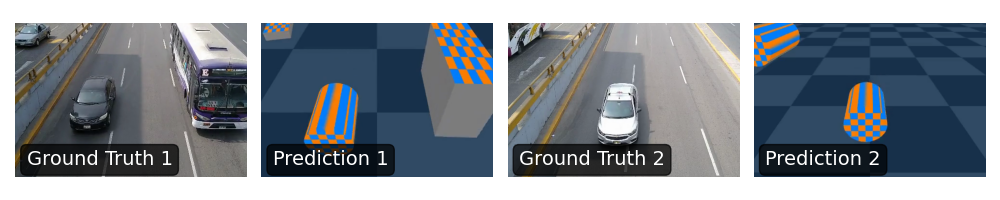}
\vspace{-9mm}
\caption{\ours{} reconstructs vehicle \textit{dynamics} in non-object-centric, in-the-wild scenes using primitive geometry.}
\vspace{-4mm}
\label{fig:highway}
\end{figure}

We show in-the-wild vehicle dynamics from DEXRAC (Figure~\ref{fig:highway}), where \ours{} reconstructs vehicle motions using primitive geometry.

\vspace{-4mm}

% \vspace{0.5em}
\subsection{Future Work} 

While our results are promising in capturing rigid-body motions using a language-centric, VLM-based approach, we identify several directions for future work: (1) incorporating 3D shape tokens~\citep{sahoo2025aligning} to move beyond primitive shapes, (2) extending to articulated objects~\citep{le2024articulate} and sloped environments to cover more types of rigid-body motion, and (3) adopting more powerful engines such as Genesis~\citep{zhou2024genesis} to model deformable objects and motions.

\section{Physically Plausible Editing}
\label{app:editing}

Our framework naturally supports physically consistent editing of object dynamics and scene parameters. Because \ours~represents each scene using an explicit and interpretable YAML configuration, user-provided editing instructions can be translated directly into modified physical parameters. This enables a closed-loop editing system that integrates a physics engine, a language model, and a video synthesis model to generate physically plausible edited videos.

\subsection{Editing Pipeline Overview}

\begin{figure*}[t]
\centering
\includegraphics[width=0.75\textwidth]{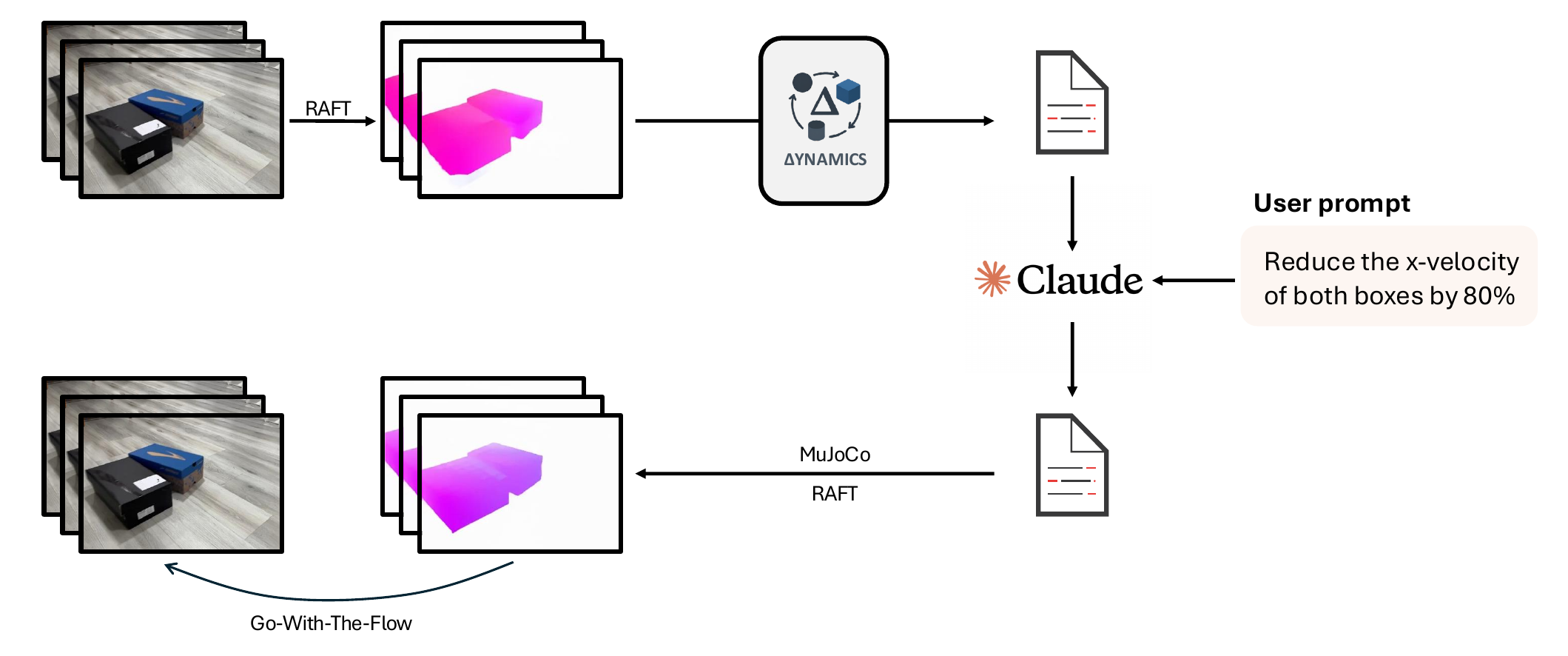}
\vspace{-2mm}
\caption{
\textbf{Physics Editing Pipeline.}
Given a user-provided editing instruction (e.g., “reduce the x-velocity by 80\%”), we first infer the original scene configuration using \ours.
Next, we prompt a large language model (Claude) with both the predicted configuration and the user instruction to generate a revised, physically consistent configuration.
The edited configuration is then executed in MuJoCo to produce a modified motion trajectory, from which we compute RAFT optical flow.
Finally, we feed the edited flow fields to Go-With-The-Flow~\citep{burgert2025go} to synthesize the edited video.
}
\vspace{-2mm}
\label{fig:supp_editing_pipeline}
\end{figure*}

Figure~\ref{fig:supp_editing_pipeline} illustrates our four-stage editing pipeline:

\begin{itemize}
\item \textbf{Configuration Extraction with \ours.}
Given an input video, we first infer its underlying physical configuration using \ours. The model outputs a complete YAML file specifying object geometries, initial poses, velocities, masses, gravity, friction, and other physical attributes. This YAML file serves as an editable and fully interpretable \textit{source code} for the scene.

\item \textbf{Language-Guided Configuration Editing.}  
To incorporate a user instruction (e.g., “reduce the x-velocity by 80\%’’ or “decrease gravity by 50\%’’), we prompt Claude-3-Haiku with (i) the full YAML configuration predicted by \ours, and (ii) the editing instruction.  
Claude outputs a revised YAML file with localized and semantically appropriate modifications (e.g., updating only the fields for initial velocity, gravity, or angular velocity).  
Because YAML is structured, line-addressable, and semantically meaningful, the language model reliably edits only the intended parameters while leaving the rest of the configuration intact. This enables precise and controllable manipulation of physical properties that would otherwise be entangled in a latent space.

\item \textbf{Simulation and Flow Generation.}  
The edited configuration is executed in MuJoCo, producing a modified motion trajectory consistent with the user’s edit. We then compute dense optical flow using RAFT~\citep{teed2020raft}, yielding a physically grounded flow field that encodes the new dynamics.

\item \textbf{Video Synthesis via Go-With-The-Flow.}  
Following \citet{burgert2025go}, the Go-With-The-Flow model synthesizes the edited video by warping noise according to the edited optical flow, conditioned on the first RGB frame of the original video. This preserves the appearance of the scene while enforcing the motion cues determined by the edited flow.

\end{itemize}

\begin{figure*}[thbp]
    \centering
    \begin{minipage}[b]{0.65\textwidth}
        \centering
        \includegraphics[width=\textwidth]{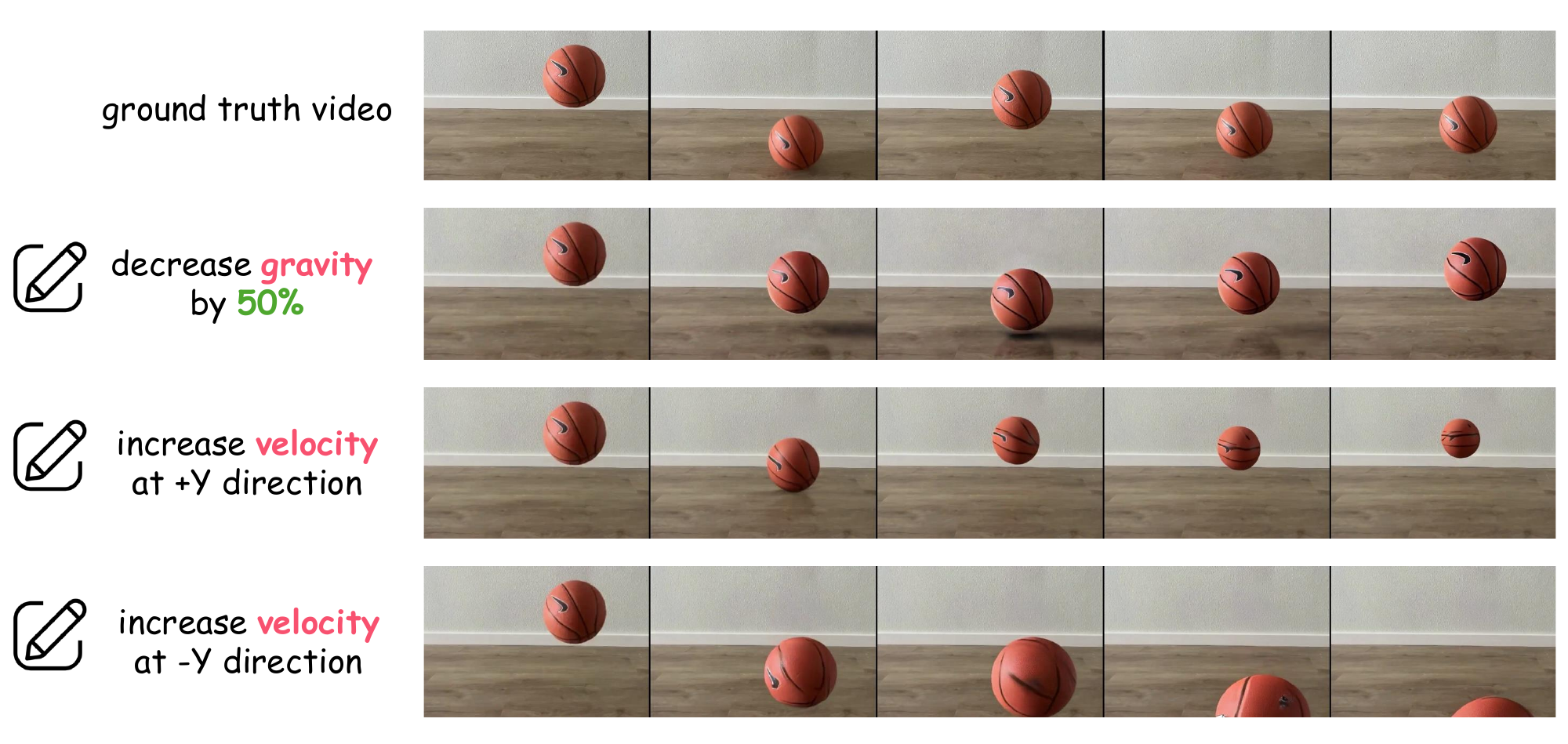}
        \caption*{
\textbf{(a) Ball Motion Editing.}
Example edits include decreasing gravity by 50\% and modifying velocity magnitude or direction.
These edits are reflected in the regenerated simulated trajectories and corresponding edited videos.
Here, positive $y$ motion indicates movement \textit{away from the camera} (deeper into the scene), while negative $y$ motion brings the object \textit{closer toward the viewer}.
}
    \end{minipage}
    \hfill
    \begin{minipage}[b]{0.65\textwidth}
        \centering
        \includegraphics[width=\textwidth]{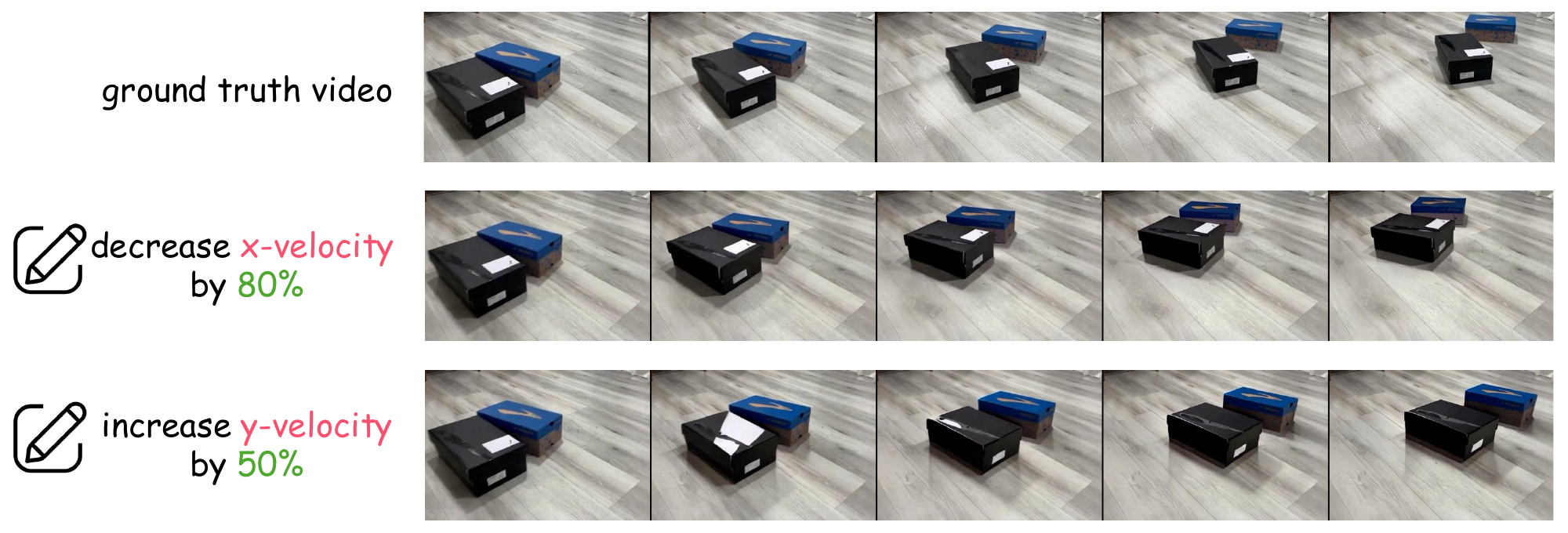}
        \caption*{
\textbf{(b) Box Collision Editing.}
Here we apply velocity reductions along specified axes (e.g., reducing x-velocity by 80\% or y-velocity by 50\%).
The resulting interactions follow physically plausible adjustments in motion and contact behavior.
}
    \end{minipage}
    \caption{CLEVRER Dataset Results.}
    \label{fig:supp_editing_examples}
\end{figure*}

Empirical results are shown in Fig.~\ref{fig:supp_editing_examples}, which demonstrates edits to both box dynamics (e.g., reducing x- or y-velocities) and ball dynamics (e.g., modifying velocity direction or reducing gravity). The pipeline produces physically correct motion and high-quality visual results in most cases.

A primary limitation arises from \textbf{appearance preservation under complex motion}.
Although Go-With-The-Flow accurately follows the edited optical flow, it sometimes struggles with fine-grained dynamic appearance details, e.g., maintaining the texture fidelity of a spinning or rolling basketball. As this remains an open challenge, future work could explore fine-tuning the generative model conditioning on edited optical flow fields to achieve better physically plausible video editing results.

\end{document}